\documentclass[10pt,twocolumn,letterpaper]{article}

\usepackage[pagenumbers,algorithms]{wacv}      

\usepackage{graphicx}
\usepackage{amsmath}
\usepackage{amssymb}
\usepackage{booktabs}
\usepackage{subcaption}
\usepackage{tikz}
\usepackage{comment}
\usepackage{amsmath,amssymb} 
\usepackage{color}
\usepackage{multirow}
\usepackage{colortbl}
\usepackage{enumitem}

\usepackage{graphicx}
\usepackage{wrapfig}

\usepackage{pifont}
\usepackage{appendix}
\usepackage{xcolor}
\definecolor{darkred}{rgb}{0.6, 0.0, 0.0}
\definecolor{darkgreen}{rgb}{0.0, 0.5, 0.0}
\newcommand{\cmark}{\textcolor{darkgreen}{\ding{51}}}%
\newcommand{\xmark}{\textcolor{darkred}{\ding{55}}}%
\usepackage[accsupp]{axessibility} 

\newcommand{\bikeemoji}{\includegraphics[height=1.5\fontcharht\font`B]{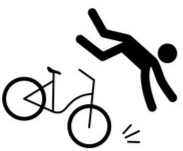}}

\usepackage[pagebackref,breaklinks,colorlinks]{hyperref}

\usepackage[capitalize]{cleveref}
\crefname{section}{Sec.}{Secs.}
\Crefname{section}{Section}{Sections}
\Crefname{table}{Table}{Tables}
\crefname{table}{Tab.}{Tabs.}

\def\wacvPaperID{447} 
\def\confName{WACV}
\def\confYear{2025}

\begin{document}

\title{
\bikeemoji{} 
CycleCrash: A Dataset of Bicycle Collision Videos for Collision\\Prediction and Analysis}

\author{Nishq Poorav Desai
~~~~~~~~~
Ali Etemad
~~~~~~~~~
Michael Greenspan\\
Queen's University, Canada\\
{\tt\small \{n.desai, ali.etemad, michael.greenspan\}@queensu.ca}
}

\maketitle

\begin{abstract}
   Self-driving research often underrepresents cyclist collisions and safety. To address this, we present CycleCrash, a novel dataset consisting of 3,000 dashcam videos with 436,347 frames that capture cyclists in a range of critical situations, from collisions to safe interactions. This dataset enables 9 different cyclist collision prediction and classification tasks focusing on potentially hazardous conditions for cyclists and is annotated with collision-related, cyclist-related, and scene-related labels. Next, we propose VidNeXt, a novel method that leverages a ConvNeXt spatial encoder and a non-stationary transformer to capture the temporal dynamics of videos for the tasks defined in our dataset.  To demonstrate the effectiveness of our method and create additional baselines on CycleCrash, we apply and compare 7 models along with a detailed ablation. We release the dataset and code 
   at \url{https://github.com/DeSinister/CycleCrash/}.
\end{abstract}

\section{Introduction}
\label{sec:intro}
The growing popularity of cycling as a sustainable and healthy means of urban commuting brings inherent risks and safety concerns placing cyclists among the most vulnerable road users~\cite{kaya2021hey,strauss2015mapping}. Recent studies show that over 130,000 cyclist injuries occur annually due to crashes, while cyclist fatalities have grown by more than 50\% over a decade \cite{cdc2022,iihs2023}. While machine learning has provided a means to address many problems related to self-driving vehicles, the development of such data-driven solutions for cyclist safety is hindered by the lack of available data that is targetted for this particular problem~\cite{isaksson2012study,10227352}. 

Some recent self-driving vehicle datasets~\cite{8639160,8967556,herzig2019spatio,yao2020and} have considered bicycles alongside cars for tasks such as object recognition, time-to-collision, and traffic scene understanding. However, the limited representation of cyclist-related instances in these datasets poses a challenge, as they make up only a small proportion of the data. For example, among widely used urban driving datasets, DoTA~\cite{yao2020and} contains only 1,031 frames with at least one cyclist, while KITTI~\cite{6248074} has only 1,627 such frames.
Moreover, while datasets like SUTD-TrafficQA~\cite{xu2021sutd} explore causal considerations like fault, the existing datasets do not explicitly provide the necessary annotations to facilitate cyclist safety, such as cyclist behaviour risk levels or collision severity. 

\begin{figure}[t]
  \center
    \includegraphics[width=0.9\columnwidth]{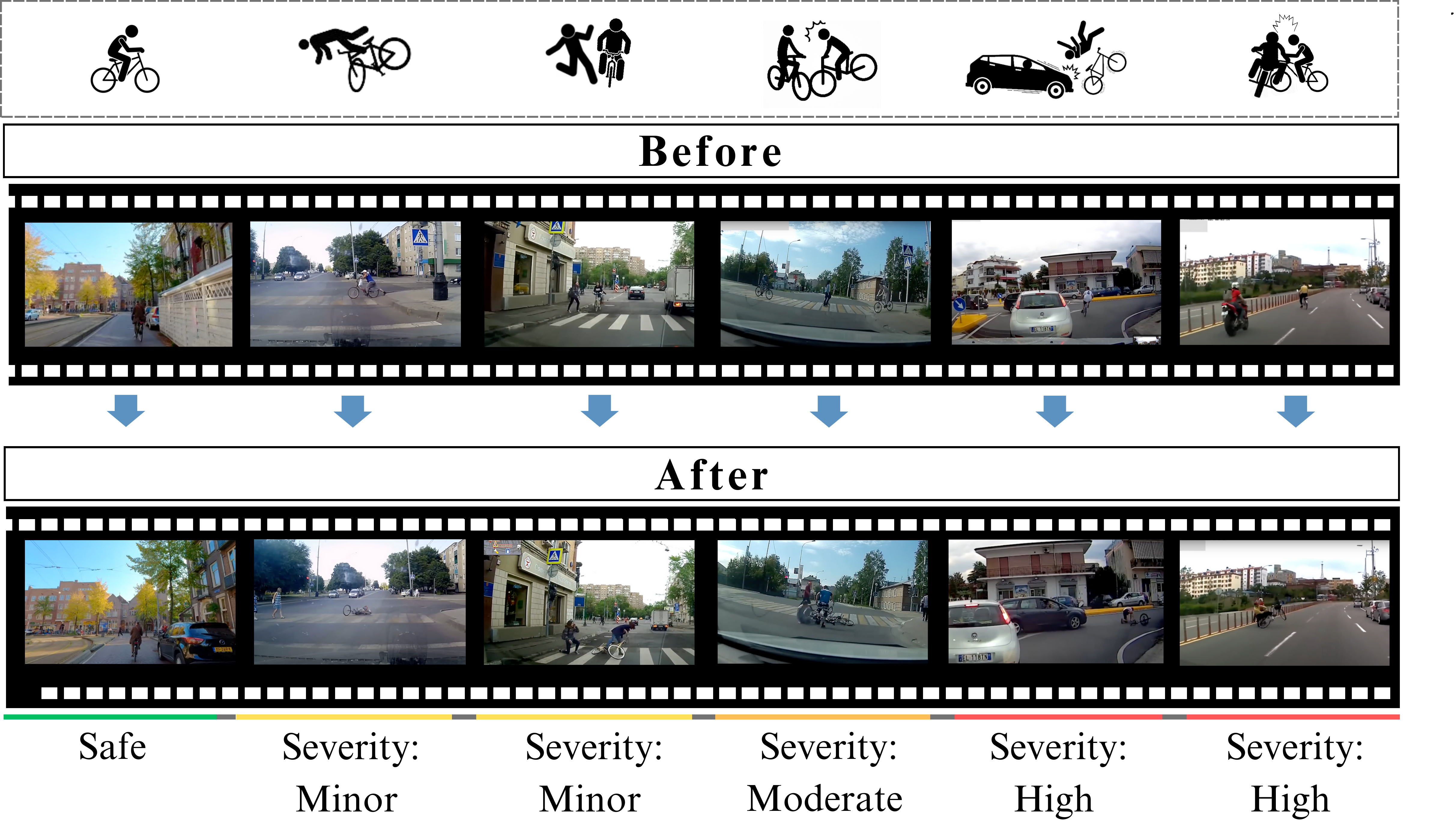}
  \caption{A few samples from CycleCrash showcasing various cyclist-related interactions with different vehicles along with collision severity levels. }
  \label{fig:graphical_abstract}
\end{figure}

\begin{table*}[h]
\center
\scalebox{0.71}{
\renewcommand\arraystretch{1}
\setlength\tabcolsep{2pt}
\begin{tabular}{lccc>{\columncolor{blue!5}}cccccccccccc}
  \toprule

  \multicolumn{1}{c}{ \rotatebox{0}{Dataset}} & 
  \rotatebox{0}{Year} & 
  \rotatebox{0}{Duration (hr)} & 
  \rotatebox{0}{Clips} &  
  \rotatebox{0}{Cyclist Frames} & 
  \rotatebox{0}{Cyclist BBox} & 
  \hspace{0pt}\rotatebox{0}{Collision Time} & 
  \hspace{0pt}\rotatebox{0}{Cyclist Age} & 
  \rotatebox{0}{Risky Behav.}& 
  \hspace{0pt}\rotatebox{0}{Severity} & 
  \rotatebox{0}{Fault} & 
  \rotatebox{0}{Right-of-way} &
  \rotatebox{0}{Cyclist Dir.} \\
\midrule
  
  KITTI~\cite{6248074} & 2012  & 1.5 & 22  & 1,627 & \cmark  & \xmark & \xmark & \xmark & \xmark & \xmark & \xmark & \xmark \\ 
  DAD ~\cite{chan2017anticipating} & 2017 & 2.4 & 1,730  & \xmark & \xmark & \xmark & \xmark & \xmark & \xmark & \xmark & \xmark & \xmark\\
  CCV~\cite{8786306} & 2018 & 0.1 & 177  & \xmark & \cmark & \cmark & \xmark & \xmark & \xmark & \xmark & \xmark& \xmark \\
  CADP~\cite{8639160}& 2018 & 5.2 & 1,416 & \(<\)50k & \cmark & \cmark & \xmark & \xmark & \xmark & \xmark & \xmark& \xmark  \\
  DACACD~\cite{8936124} & 2019 & 5.0 & 392 & \xmark & \cmark & \cmark & \xmark& \xmark & \xmark & \xmark & \xmark & \xmark \\
  A3D~\cite{8967556} & 2019 & 3.6 & 1,500 & \xmark & \cmark & \cmark & \xmark & \xmark & \xmark & \xmark & \xmark & \xmark \\ 
  STAG-Nets~\cite{herzig2019spatio} & 2019 & 8.9 & 803 & \xmark  & \xmark & \cmark & \xmark & \xmark & \xmark & \xmark & \xmark & \xmark \\
  BDD100k~\cite{bdd100k} & 2020  & 1,111.1 & 100k & 13,122\(\dagger\)  & \cmark & \xmark & \xmark & \xmark & \xmark & \xmark & \xmark & \xmark \\
  DoTA~\cite{yao2020and} & 2020  & 20.0 & 4,677 & 1,031 & \cmark & \cmark & \xmark & \xmark & \xmark & \cmark & \xmark & \xmark \\ 
  WOD~\cite{sun2020scalability} & 2020  & 10 & 1,000 & \xmark  & \cmark & \xmark & \xmark & \xmark & \xmark & \xmark & \xmark & \xmark \\  
  nuScene~\cite{caesar2020nuscenes} & 2020  & 5.5 & 1,000  & \(<\)11.8k &  \cmark & \xmark & \xmark & \xmark & \xmark & \xmark & \xmark  & \xmark\\ 
 CCD~\cite{10.1145/3394171.3413827} & 2020 & 6.3 & 1,500 & \xmark & \xmark & \cmark & \xmark & \xmark & \xmark & \xmark & \xmark & \xmark\\
  RetroTrucks~\cite{9304576} & 2020 & 1.9 & 474 &  \xmark & \cmark & \cmark & \xmark & \xmark & \xmark & \xmark & \xmark & \xmark \\ 
  DADA2000~\cite{fang2021dada} & 2021 & 6.1 & 2,000 & \xmark & \xmark & \xmark & \xmark& \xmark & \xmark & \xmark & \xmark& \xmark \\ 
  SUTD-TrafficQA~\cite{xu2021sutd}& 2021 & \xmark & 10k &  \xmark & \cmark & \cmark & \xmark & \xmark & \xmark & \cmark & \xmark & \xmark \\
  Lyft Level5~\cite{houston2021one} & 2021  & 1,118 & 170k & 20,928\(\dagger\)  & \cmark & \xmark & \xmark & \xmark & \xmark & \xmark & \xmark & \xmark\\
  \hline
  \textbf{CycleCrash (ours)} & 2024  & 4.1 & 3,000 & 436k  & \cmark & \cmark & \cmark & \cmark & \cmark & \cmark & \cmark& \cmark  \\
  \bottomrule
\end{tabular}
}
\caption{Comparison of CycleCrash against existing datasets. $\dagger$ indicates partial annotation, 
\cmark ~indicates that the label is available, while \xmark ~indicates that the information/label is unavailable or unknown.}
\vspace{-1mm}
\label{tab:dataset_comparison}
\end{table*}

To address this issue and facilitate deep learning solutions for cyclist safety, we present \textbf{CycleCrash}, a novel dataset that explicitly addresses several key challenges in this area. 
CycleCrash comprises $3,000$ videos sourced primarily from the web.
In summary, we make the following contributions.
\noindent (\textbf{1}) 
CycleCrash comprises 3,000 video clips of cyclist-related scenes along with \textbf{13 types of annotation}, organized into the 3 categories of \emph{collision-related}, \emph{cyclist-related}, and \emph{scene-related}. \cref{fig:graphical_abstract} shows different interactions of cyclists in the CycleCrash dataset along with associated collision severity levels.
(\textbf{2}) 
We define \textbf{9 key tasks} (classification or regression) based on our dataset, with implications on cyclist safety.
(\textbf{3}) 
To accurately perform the defined safety-related tasks based on the collected data, we propose VidNeXt, a novel architecture for video representation learning. VidNeXt combines a ConvNeXt feature extractor with a non-stationary transformer for the first time and is, explicitly designed to learn both stationary and non-stationary spatiotemporal information from videos, which allows it to perform accurate prediction and classification. 
Through extensive experiments, we show that our model outperforms various baselines on most CycleCrash tasks.

\section{Related Works}\label{sec:relatedwork}
We summarize existing crash-related datasets from the literature in \cref{tab:dataset_comparison}, and observe that none specifically focus on cyclists. Therefore in this section, we present the next most relevant body of work, i.e., datasets and methods focusing on self-driving vehicles, and car collisions.

\noindent \textbf{Car Collision Datasets.}
Early datasets focusing on car collisions, such as Car Crash Videos (CCV)~\cite{8786306} and its successors~\cite{sultani2018real,8639160,8936124,xu2021sutd}, include annotations of car collisions viewed from the vantage of fixed perspective cameras such as pole-mounted CCTV cameras. In contrast,  newer datasets use dashcam videos widely accessible from internet sources. 
As observed in \cref{tab:dataset_comparison}, CADP~\cite{8639160} offers the highest number of cyclist frames among the datasets that report this count, with fewer than 50,000 frames.
This demonstrates a clear gap, as it encompasses only about one-ninth of the cyclist frames available in the CycleCrash dataset. 

Additionally, it is observed that despite containing 3,000 video clips, CycleCrash's total duration is only 4.1 hours, which is shorter than some other datasets containing fewer video clips.
This is expected because we temporally crop the video clips in CycleCrash to exclude those parts of the video which do not contain relevant cyclist information.

A wide variety of methods for addressing traffic safety have been previously outlined, such as identifying the occurrence of collisions which was first introduced by the DAD dataset~\cite{chan2017anticipating}, `time-to-collision' which was coined by the CADP dataset~\cite{8639160}, as well as SUTD-TrafficQA dataset~\cite{xu2021sutd}, which is a video-based question-answering dataset that also includes causal considerations such as `fault'. However, annotations such as `right-of-way' have been relatively unexplored. CycleCrash introduces novel annotations such as the `cyclist behaviour risk index' and `severity', which have not been utilised before to the best of our knowledge. These annotations can play an important role in responding promptly to avoid collisions, as well as potentially alerting first responders in critical situations.
Furthermore, the types of annotations in existing datasets are often fragmented, highlighting the absence of a unified dataset with credible cyclist frames and a comprehensive collection of annotations for cyclist safety.
Owing to its specific cyclist-related focus, CycleCrash provides a unique resource to
enable increased cyclist safety, e.g., cyclist accident prediction, time-to-collision forecasting, \etc., for the autonomous driving and driver assist research community. \\
\noindent \textbf{Methods.}
The evolution of urban-traffic-related video classification and detection methods closely aligns with dataset diversity. Early approaches focused on conventional methods such as YOLO~\cite{bochkovskiy2020yolov4} for binary collision detection~\cite{8786306}. Convolutional neural nets (CNNs) when in combination with LSTMs~\cite{9481040}, or in the form of 3D-CNN (such as ResNet3D, X3D, DensNet3D \etc)~\cite{hara2018can,feichtenhofer2020x3d} show prominent application in urban traffic video-based tasks~\cite{zhang2019vehicle,muthuswamy2023driver,yang2021crossing,anjum2023spatio}, although they are somewhat computationally expensive.
R(2+1)D~\cite{tran2018closer} was introduced to address these computational drawbacks with divided spatial and temporal CNNs. 


Transformers~\cite{vaswani2017attention} have demonstrated tremendous success in the vision domain, and have been adapted for urban-traffic-related video tasks including object detection, multiple object tracking~\cite{huang2023video}, natural driving action recognition~\cite{dong2023multi} and anomaly detection~\cite{chang2022video,doshi2023towards,liu2023learning,9263703,akdag2023transformer}
. Examples of such transformer-based methods include ViViT~\cite{arnab2021vivit}, VideoMAE~\cite{tong2022videomae,wang2023videomae} and TimeSformer~\cite{gberta_2021_ICML}, all of which also apply similar spatiotemporal division potentially inspired by R(2+1)D. However, they still are ineffective in capturing global representation and aren't specifically designed to handle stationary and non-stationary elements of temporal information.
In processing the temporal domain,  a common pre-processing step is to adopt stationarization to improve predictability, albeit at the cost of losing non-stationary information. We propose VidNeXt to address this issue by using both stationary and non-stationary components for improved video representation learning (see  \cref{sec:proposed_method}).

\section{CycleCrash Dataset}\label{sec:cyclecrash}

To address the limitations identified in prior works regarding cyclist safety, we introduce CycleCrash. In this section, we describe the collection procedure, data format, annotations, and other details of 
this dataset. The dataset along with the 
tool kit (see Appendix A for  details)
will be made publicly available upon publication of the work.

\begin{table*}[h]
\centering
\scalebox{0.7}{
\renewcommand\arraystretch{1}
\centering
\setlength\tabcolsep{6pt}
\begin{tabular}{|c|p{4.5cm}|p{7.5cm}|p{6.5cm}|}
\hline
\textbf{No.} & \textbf{Annotation} & \textbf{Description} & \textbf{Options} \\
\hline\hline
1 & Right-of-way & Binary label capturing the priority between the cyclist of interest and others involved (e.g., vehicle, other cyclists, pedestrian) & Yes, No \\
\hline
2 & Time-to-collision & Exact timestamp on which the collision begins if one exists in the video clip & Timestamp (if a collision occurs) \\
\hline
3 & Type of object involved & 17-class categorization to describe the type of object involved in the interaction & Car, Bus, Train, Cyclist, Pedestrian, Pothole, Animal, etc. \\
\hline
4 & Fault & Binary label determining whether the collision was the cyclist of interest's fault & Yes, No \\
\hline
5 & Severity & 5-class categorization to describe the intensity of the collision & Safe, Minor, Moderate, High, Very high \\
\hline
6 & Cyclist behaviour risk index & 
4-class categorization to indicate the risk levels portrayed by the cyclist
& Low, Moderate, High, Very High \\
\hline
7 & Cyclist age & 3-class categorization of the age of the cyclist. & Young, Adult, Old \\
\hline
8 & Cyclist type & Binary label capturing if the cyclist is participating in competitive or recreational cycling & Competitive, Recreational \\
\hline
9 & Cyclist bounding box & Bounding box captured around the interested cyclist & Coordinates in $x$,$y$,$w$,$h$ format \\
\hline
10 & Direction of the cyclist & 5-class categorization of the direction of the cyclist's motion as observed from the ego-vehicle along the medial and lateral directions & Forward (+ve medial motion $x+$), Backward (-ve medial motion $x-$), Left (-ve lateral motion $y-$), Right (+ve lateral motion $x+$), and stationary (no motion) \\
\hline
11 & Direction of the object involved & 5-class categorization of the direction of the object involved as observed from the ego-vehicle along the medial and lateral directions & Forward, Backward, Left, Right, Stationary \\
\hline
12 & Camera position & Categorization based on the camera's position (Front/Back) and the camera carrying vehicle & Front dashcam, Back dashcam, Front cyclist-helmet camera etc. \\
\hline
13 & Ego-vehicle involved & Binary label indicating whether the ego-vehicle is involved in a collision or not & Yes, No \\
\hline
\end{tabular}
}
\caption{Descriptions of annotations provided in CycleCrash along with possible options.}
\label{tab:annotations}
\end{table*}

\subsection{Data Collection}\label{subsec:data_collection}
We curated the dataset from the web, namely from  
YouTube \cite{youtube}, Vimeo \cite{vimeo}, DailyMotion \cite{dailymotion}, Facebook \cite{facebook}, Instagram \cite{instagram}, X (formerly known as Twitter) \cite{x}, and TikTok \cite{tiktok}. Each video contains one or more cyclists, as viewed from vehicle dashcams. The following are the criteria used to qualify a video for inclusion in our dataset, where the first four constitute a collision or a near-miss:
\begin{enumerate}[label=\textbullet, nosep]
    \item There exists a collision or near-miss involving a cyclist and a motor vehicle (i.e. car, motorcycle, bus, etc.);
    \item There exists a collision between a cyclist and a pedestrian or another cyclist;
    \item There exists a loss of balance or fall due to 
    potholes, animals, mechanical failure of the bicycle, etc.; 
    \item There exists risky behaviour shown by the cyclist, justifying the need for extra attention to be given by drivers to prevent a collision;
    \item There exists a cyclist navigating safely in an urban traffic scenario, without depicting risky behaviour or being involved in a collision or near-miss.
\end{enumerate}
The dataset comprises 3,000 video clips, out of which 2,000 are accident-free and 1,000 exhibit potential accidents of various levels of severity and near-misses involving a cyclist of interest. We collected dashcam videos instead of fixed-perspective CCTV footage to facilitate the development and integration of cyclist collision warning and mitigation systems in vehicles.

\subsection{Data Format and Structure}\label{subsec:data_format}

CycleCrash is a curated list of video links accompanied by precise start and end timestamps. 
All videos are sourced explicitly from public posts, ensuring no private videos are used. Following established practices from prior works, such as \cite{10.1145/3394171.3413827,yao2020and,8967556}, we do not download or share the videos. Instead, we provide links to the original videos along with the relevant timestamps.
We create and include a library with our dataset, that downloads and pre-processes the videos for uniformity in spatiotemporal properties. Our pre-processing pipeline includes: 

\noindent \textbf{Temporal Cropping.} 
Each video is temporally cropped based on the provided start and end times.
 
\noindent \textbf{File Conversion.} 
All videos are converted to .mp4 format.

\noindent \textbf{Spatial Cropping and Scaling.}
Videos are rescaled to a uniform resolution of $1,280 \times 720$ pixels using interpolation. When a video is not in a $16/9$ aspect ratio, they are cropped before rescaling to achieve the desired aspect ratio. 

\noindent \textbf{Frame Rate Adjustment.} 
Videos are adjusted to a consistent frame rate of $30$ frames per second (fps). For videos that were initially higher than $30$ fps, our library performs temporal sub-sampling, whereas, for videos with lower fps rates, the library performs up-sampling using interpolation.

\noindent \textbf{Color Normalization.} 
Videos are z-score normalized.

\subsection{Dataset Annotation}\label{subsec:data_annot}
We conducted a comprehensive survey on possible attributes for safety~\cite{suzuki2018anticipating,9263703,kay2017kinetics}  and compiled a list, which we then narrowed down to 13 annotations based on perceived importance.
The $13$ different annotations, organized into three categories, are as follows:

\noindent \textbf{Collision-related}, which refers to characteristics associated with annotations in the context of a collision or close call. Factors like the right-of-way and time of collision contribute to enabling autonomous vehicles to predict and avoid collisions. They include `right-of-way', `time-to-collision', `type of object involved', `fault', `severity';

\noindent \textbf{Cyclist-related}, which captures information regarding the cyclist of interest e.g., cyclist behaviour risk index, age or appearance. They include `cyclist behaviour risk index', `cyclist age', `cyclist type' (competitive or recreational), `cyclist bounding box', and `direction of the cyclist';

\noindent \textbf{Scene-related}, which provides additional information regarding the motion of the vehicle involved in the scene, as well as the cameras used to capture the videos. They include the `direction of the object involved', `camera position', and `ego-vehicle involved'.

We present a detailed description of each annotation in \cref{tab:annotations}, and examples of annotated videos in \cref{fig:annotation_showcase} (additional visuals are presented in Appendix F).

\subsection{Quality Control}
The videos were collected based on search terms such as `cyclist hitting car', `bicyclist car accident', `cyclist accident dashcam', and others (please see the full list in Appendix B). The videos were automatically processed to ensure their frame rates were at least 20 fps, and that their resolutions were greater than 854 $\times$ 480 pixels. Following this, they underwent further human inspection to ensure the conditions in \cref{subsec:data_collection} were met. The selected set of videos were then preprocessed and filtered algorithmically into the common configuration described in \cref{subsec:data_format}. 

To ensure bias mitigation for subjective annotations such as risk, we provided labellers with detailed instructions to pay attention to traffic rules, legal speed, distance from other objects and vehicles, etc. For age, labellers were instructed to consider the visual appearance of the cyclist. For severity, labellers were asked to pay attention to collision impact and degree of estimated injury caused to the cyclist. When video data lacked sufficient information to confidently determine scores for `right-of-way', `fault', `age', and `cyclist type', annotators were asked to use `-1'.
Each video was annotated by three different labellers with valid driving licenses and sufficient knowledge of traffic conditions. We employed the Latin Square method~\cite{benjamin1965special} for labelling, which presented each labeller with an equal distribution of tasks and scenarios, thereby reducing bias and enhancing the quality of the dataset's annotations. For categorical annotations we use the median, while for continuous labels, an average of the values is used. Qualitative labels including `risk' and `age' have been previously labelled in other datasets ~\cite{suzuki2018anticipating,kogure2022age}, and we followed a similar labelling protocol. Lastly, we follow~\cite{schlichtkrull2024averitec,amidei2019agreement}, and use Randolph's Coefficient \cite{randolph2005free} to measure inter-annotator agreement for `cyclist age' (0.81), `severity' (0.76), and `risk' (0.84), which are considerably higher than previous works and indicate substantial agreement among labellers.

\begin{figure*}[!t]
  \centering
  \begin{subfigure}{0.74\linewidth}
    \includegraphics[width=\linewidth]{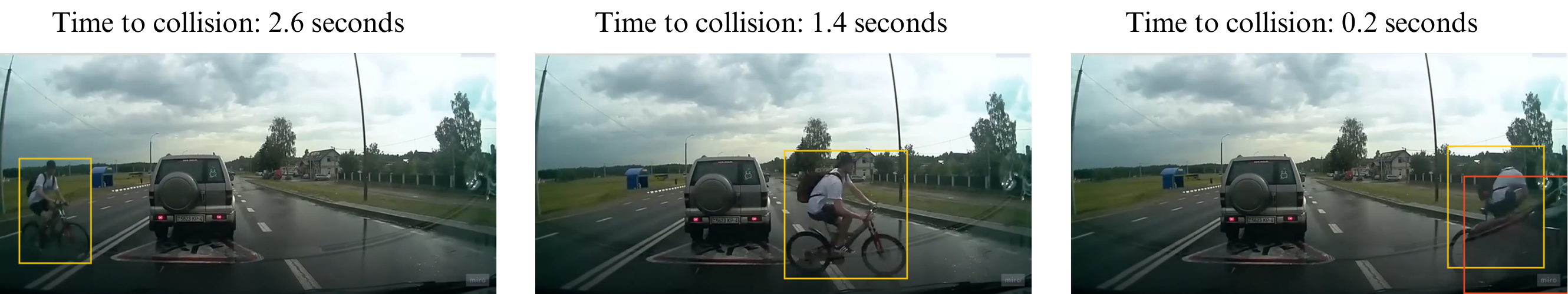}
    \caption{
    A young recreational cyclist, with a very high risk index, and a direction of left to right, is depicted. A car carrying the dashcam records another car driving forward and colliding with the cyclist
    at 2.6 seconds from the first frame. Here the car had the right-of-way and the cyclist was at fault.
    }
    \label{fig:annotation_showcase_a}
  \end{subfigure}
  \begin{subfigure}{0.74\linewidth}
    \includegraphics[width=\linewidth]{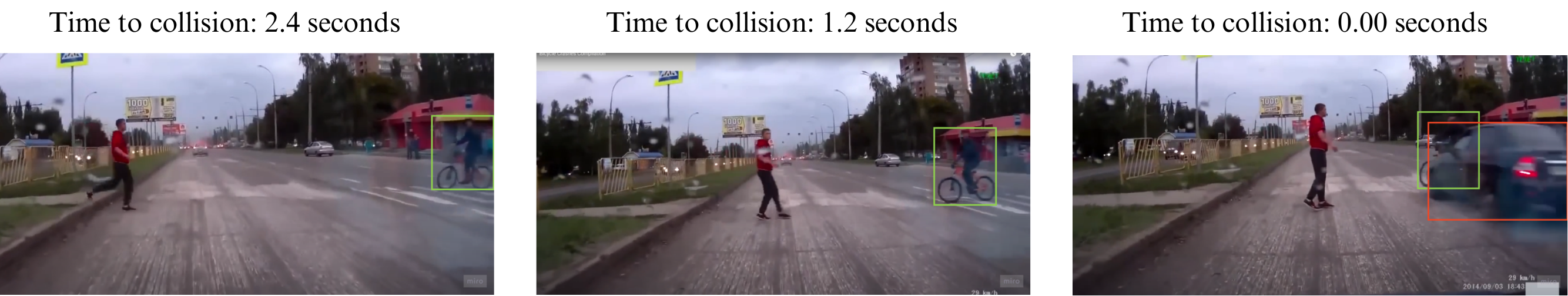}
    \caption
    {
    An adult recreational cyclist, with a low risk index, and a direction of right to left, is depicted. A dashcam on a car records another car driving forward and colliding with the cyclist
    at 2.4 seconds from the first frame. Here the cyclist had the right-of-way and the car was at fault.
   }
    \label{fig:annotation_showcase_b}
  \end{subfigure}
  \begin{subfigure}{0.74\linewidth}
    \includegraphics[width=\linewidth]{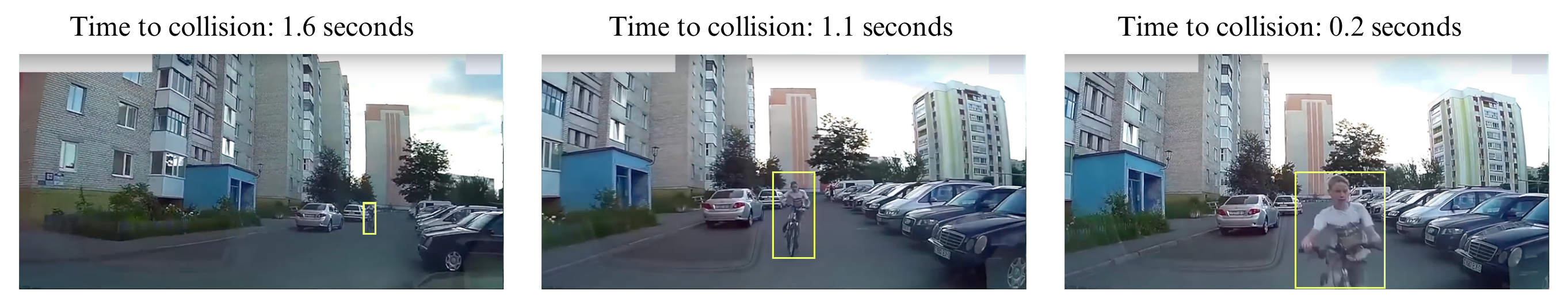}
    \caption
    {
   A young recreational cyclist, with a moderate risk index, and a backward direction, is depicted. A car carrying the dashcam drives forward and collides with the cyclist
   at 1.6 seconds from the first frame. Here the car had the right-of-way and the cyclist was at fault.
    }
    \label{fig:annotation_showcase_c}
  \end{subfigure}
  \caption{Sample frames from 3 video clips along with descriptions and annotations from the CycleCrash dataset.}
  \vspace{-2mm}
  \label{fig:annotation_showcase}
\end{figure*}


\subsection{Tasks and Metrics}\label{subsec:tasks_and_metrics}

We define $9$ tasks based on the 
labels, as follows. 

\noindent\textbf{Task 1: Cyclist Behaviour Risk Index Prediction}
estimates the cyclist's level of risky behaviour based 
collected for our dataset which is first recorded on a continuous scale of 0 to 1 and subsequently quantized into the 4 classes `Low', `Moderate', `High', and `Very High', to reduce subjective variability and cast the task as a classification problem rather than regression.
We adopt accuracy and F1 as the metrics to quantify the performance on this task.

\noindent\textbf{Task 2: Right-of-way Classification}
involves classifying the priority in a cyclist's interaction with another vehicle or pedestrian. We treat this task as a binary classification to determine if the cyclist has the right-of-way or not. 
We use accuracy and F1 to evaluate performance on this task. 
Although a few prior works~\cite{xu2021sutd,10.1145/3394171.3413827} have implicitly analyzed right-of-way as a fault-related factor, they have not exclusively outlined right-of-way as a separate task. 

\noindent\textbf{Task 3: Cyclist Collision Anticipation}  
predicts whether the sequence will result in a collision or not within a specified timeframe of $t$ seconds, where $t$ is a predefined horizon. 
Prior works~\cite{8786306,8936124,xu2021sutd,yao2020and,10.1145/3394171.3413827,9304576} have explored this task in other contexts using horizons ranging from $0.33$ to $0.64$ seconds, whereas we standardize on a horizon of 1 second.
We use binary accuracy to quantify performance on this task.

\noindent\textbf{Task 4: Time-to-Collision Prediction}
is a regression task that involves predicting the exact moment when a collision with a cyclist will occur following the completion of the video sequence. 
We use MSE for the evaluation of this task.

\noindent\textbf{Task 5: Severity Classification}
indicates the impact of an accident 
on the cyclist into 5 classes: `safe', `minor', `moderate', `high', and `very high'. 
We employ both accuracy and F1 as evaluation metrics for this task.

\noindent\textbf{Task 6: Fault Classification} 
involves characterizing the fault in the case of a collision or a near-collision between the cyclist and the other object involved. We treat this as a binary classification task to estimate whether the cyclist was at fault. 
We use accuracy and F1 to evaluate performance on this task.

\noindent\textbf{Task 7: Cyclist Age Classification} involves classifying the age of the cyclist of interest into one of the three categories: `young', `adult', or `old'. 

\noindent\textbf{Task 8: Direction of the Cyclist Detection} classifies the final direction of the cyclist in the given video into five different possible classes: `forward', `backward', `left', `right', or `stationary'. We employ both accuracy and F1 as evaluation metrics for this task.

\noindent\textbf{Task 9: Direction of the Object Involved Detection} is a similar task to `Direction of the Cyclist Detection', which aims to estimate the final direction of the other object involved with the five possible categories.

For the task of `Cyclist Behaviour Risk Index Prediction', `Direction of the Cyclist Detection', and `Cyclist Age Classification', all $3,000$ videos (both collision and safe) were considered. For the rest of the tasks, the $1,000$ collision or near-miss videos were used. 
We provide the results for a multi-task implementation with all 9 tasks in Appendix C and discuss future research directions in Appendix D.

\subsection{Data Statistics}\label{subsec:data_stats}

\begin{figure}[t]
  \centering
    \includegraphics[width=\linewidth]{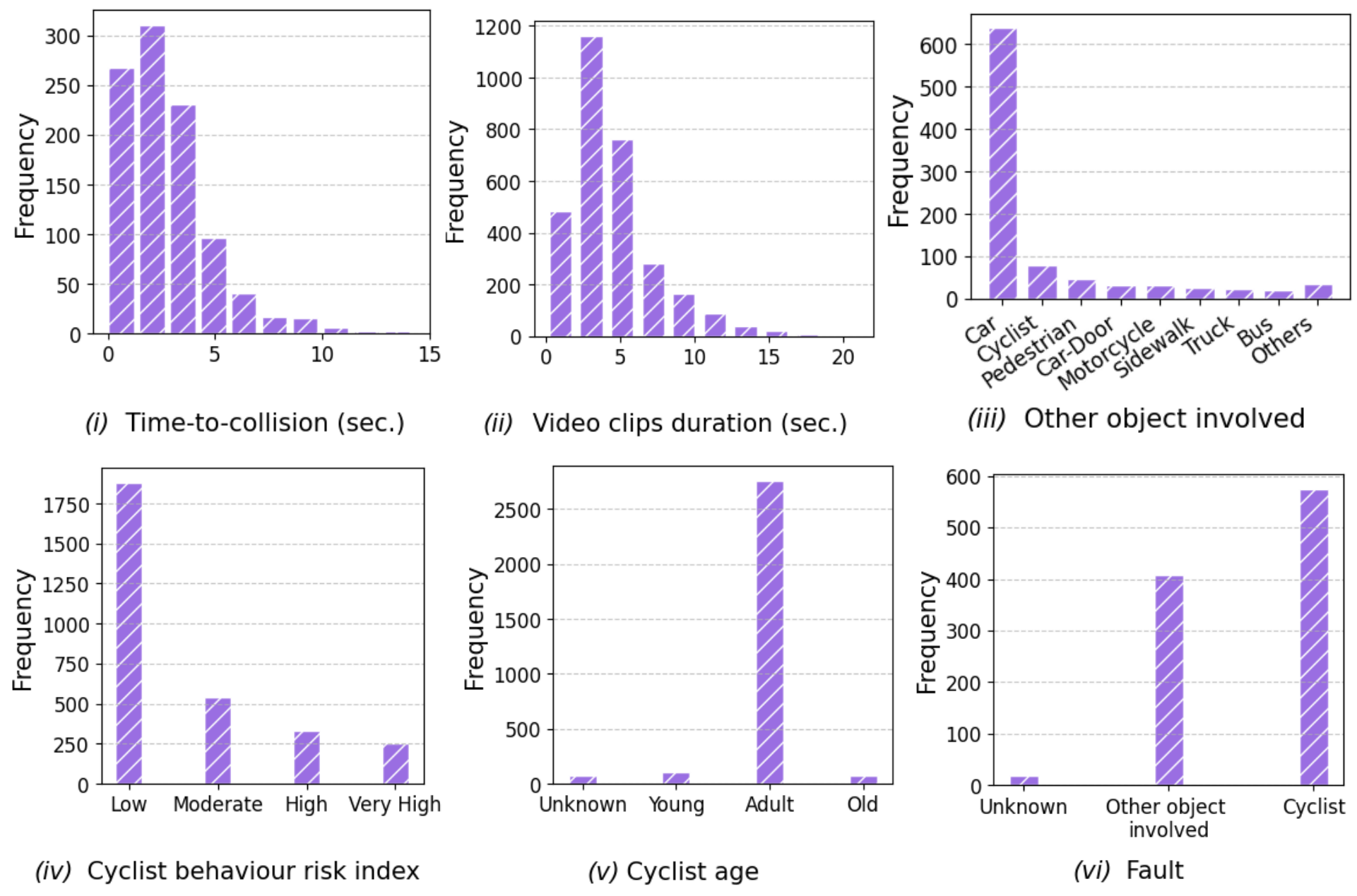}
  \caption{Distribution of CycleCrash data for (\textit{i}) time-to-collision, (\textit{ii}) duration of video clips, (\textit{iii}) other objects involved, (\textit{iv}) behaviour risk index, (\textit{v}) age, and fault.} 
\label{fig:data_stat_hist}
\end{figure}

\begin{figure}[t]
  \centering
    \includegraphics[width=1\linewidth]{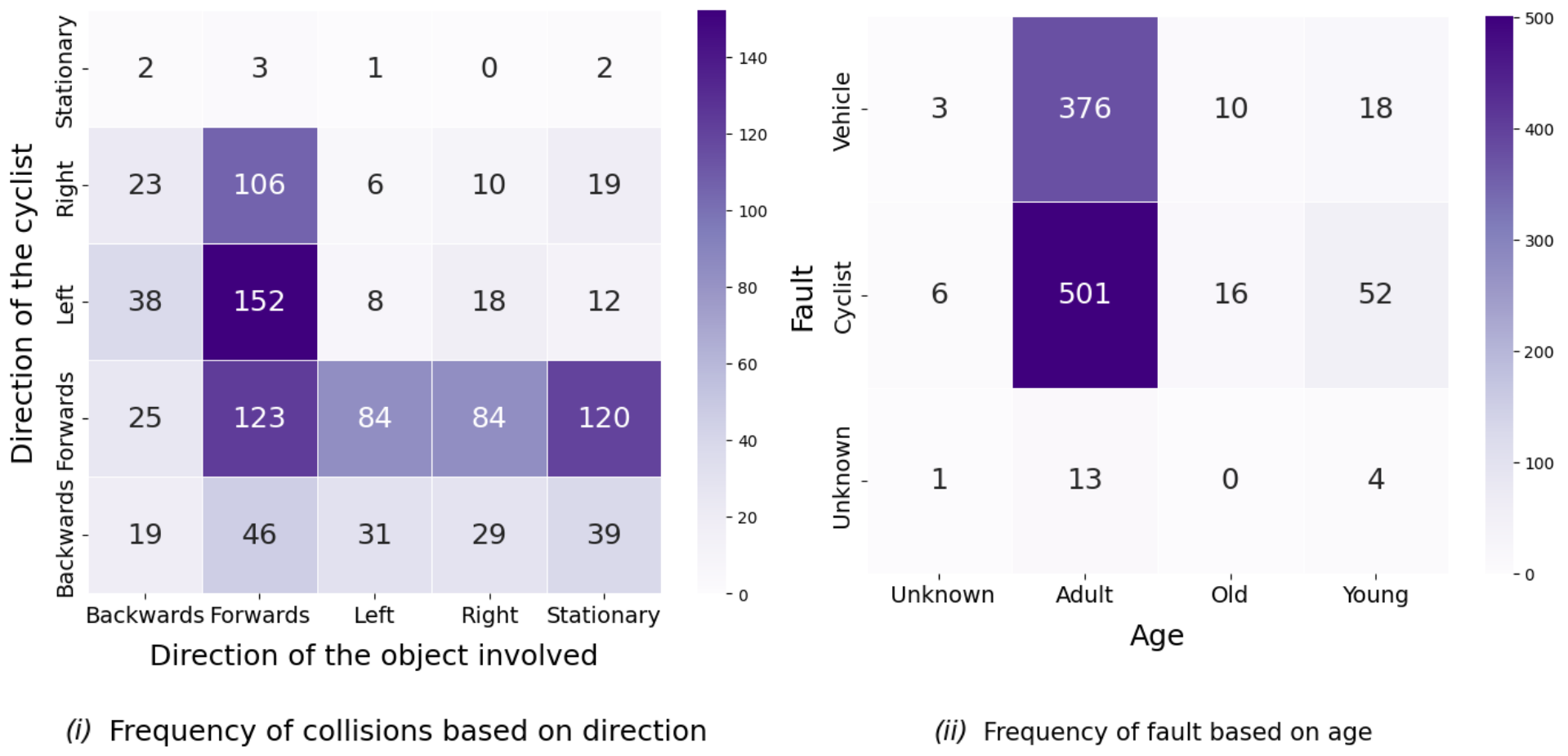}
    \caption{Relationship between (\textit{i}) direction of cyclists and objects involved in collisions, (\textit{ii}) fault and age.}
    \label{fig:data_stat_heatmap}
  \end{figure}

The proposed CycleCrash dataset contains $3,000$ videos, with durations varying from $1.5$ to $21$ seconds, totalling over $436,347$ frames. We present the key statistics of our dataset in \cref{fig:data_stat_hist} and \cref{fig:data_stat_heatmap}.  
\cref{fig:data_stat_hist} (\textit{i}) illustrates the distribution of time-to-collision in seconds, while \cref{fig:data_stat_hist} (\textit{ii}) highlights the distribution of the duration of all video clips in the dataset. \cref{fig:data_stat_hist} (\textit{iii}) provides a breakdown of the types of objects involved in cyclist collisions. We observe that in the majority of videos, the other object involved is the car as expected. \cref{fig:data_stat_hist} (\textit{iv}) presents the distribution of instances across different cyclist behaviour risk indexes, suggesting that in most cases, cyclists show reasonably low-risk behaviour across the dataset. \cref{fig:data_stat_hist} (\textit{v}) depicts the perceived age distribution of the cyclists in the videos, indicating that the majority of subjects are adults. Finally, \cref{fig:data_stat_hist} (\textit{vi}) presents the fault distribution, demonstrating that fault is reasonably similar between cyclists and others.

\cref{fig:data_stat_heatmap} (\textit{i}) displays the frequency of collisions based on the directions of the cyclist and the other objects involved relative to the ego-motion of the dashcam. It is observed that the highest number of collisions occur when the direction of the cyclist is toward the left and the direction of the other object involved is forward. Additionally, we observe that the lowest collisions occur when the cyclist is stationary. To clarify, the occurrence of rare accidents between stationary cyclists and stationary objects ($n$ = 2) refers to instances where the object, e.g., a traffic sign, falls on a stationary cyclist. Moreover, \cref{fig:data_stat_heatmap} (\textit{ii}) presents fault against cyclist age, where we observe that young cyclists are 3 times more likely to be at fault compared to adults, while this factor is 1.6 times for older cyclists. Additional statistics are provided in Appendix E.

\section{Proposed Method: VidNeXt}\label{sec:proposed_method}

\begin{figure*}[t]
  \centering
    \centering
    \begin{subfigure}{0.67\linewidth}
      \centering
      \includegraphics[width=\linewidth]{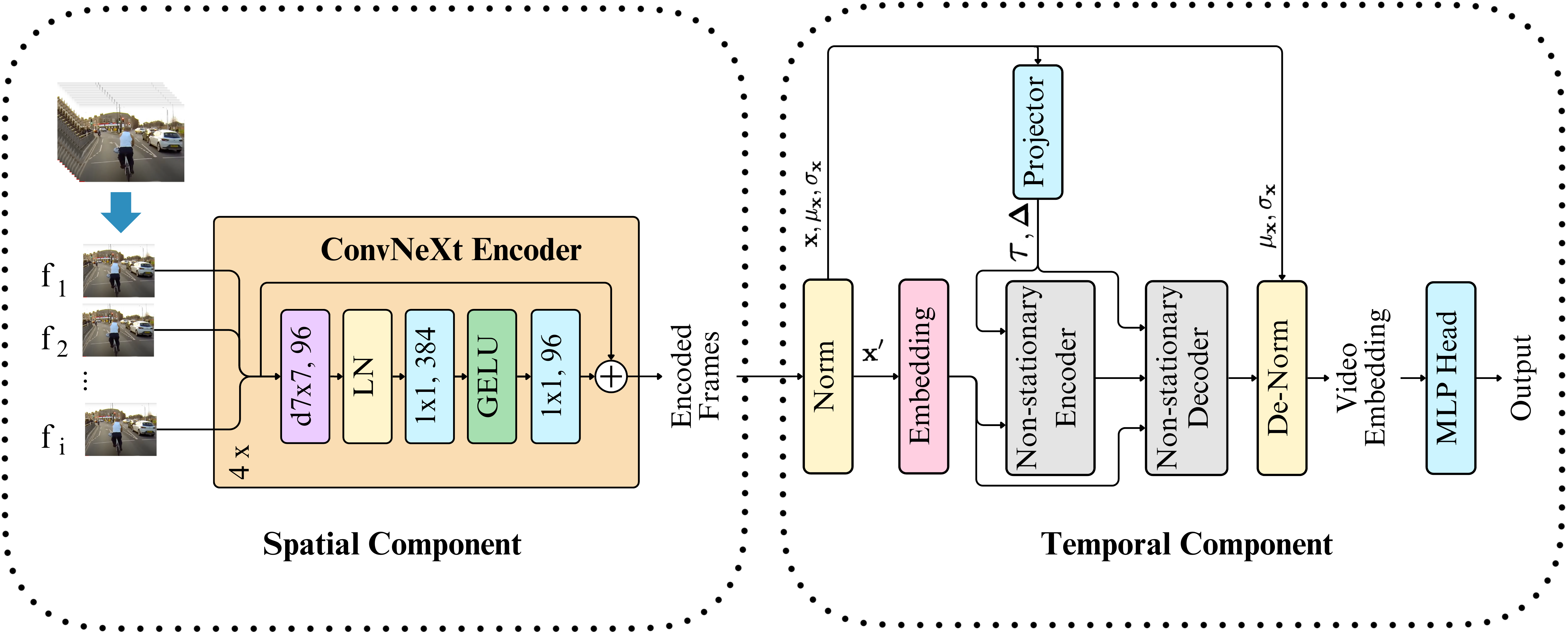}
      \caption{VidNeXt architecture}
      \label{fig:arch_diag1}
    \end{subfigure}
    \begin{subfigure}{0.32\linewidth}
      \centering
      \includegraphics[width=\linewidth]{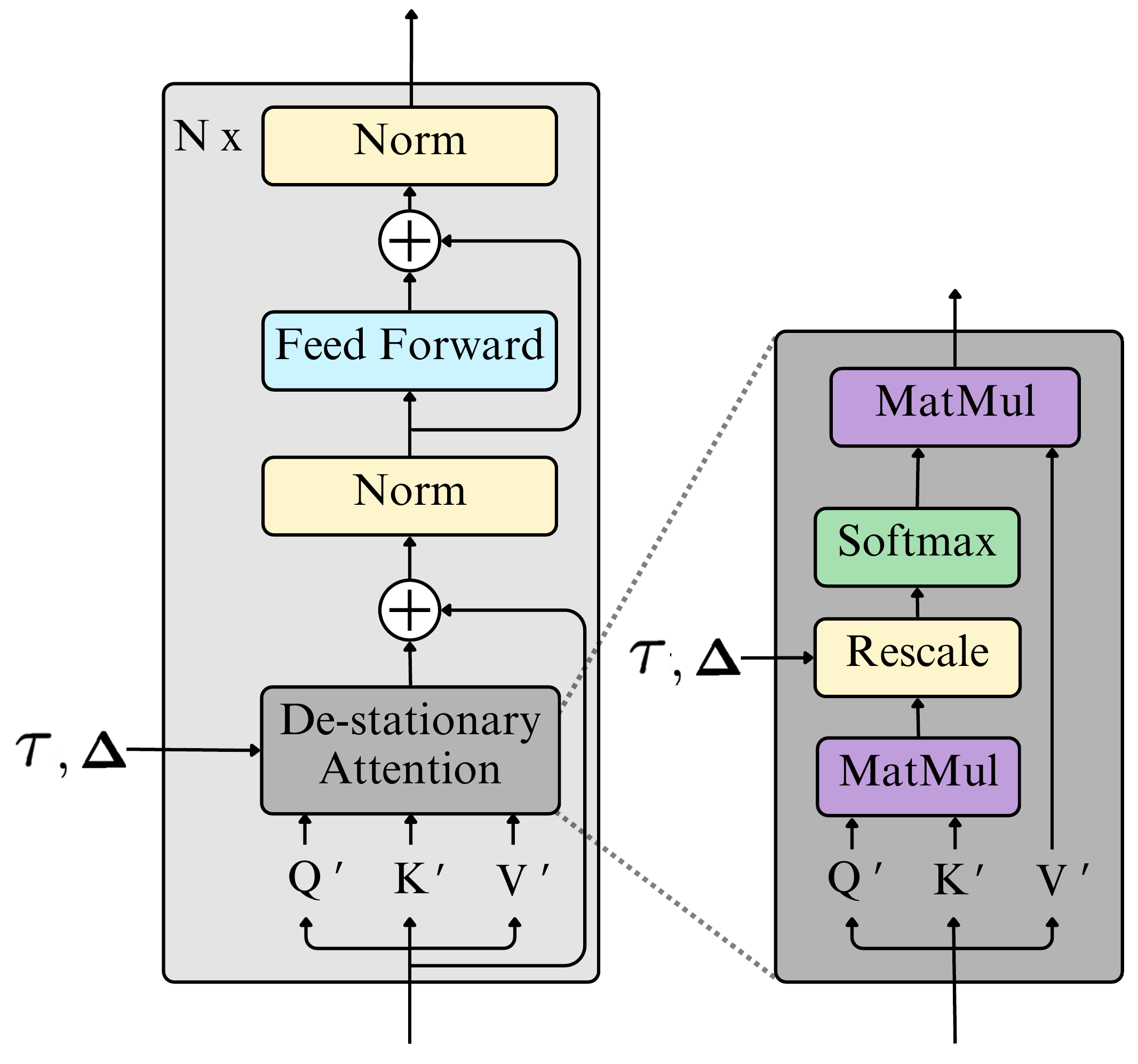}
      \caption{Transformer block with de-stationary attention}
      \label{fig:arch_diag2}
    \end{subfigure}
   
    \caption{\textbf{(a)} The architecture of the proposed method, VidNeXt is presented. First, ConvNeXt is used to encode the video frames. Next, the frame embeddings are normalized for stationarity. Non-stationary information is reintegrated into the transformer blocks via rescaling factors $\tau$ and $\Delta$, determined by the projector based on the previous normalization layer.  \textbf{(b)} Architectural details of transformer with de-stationary attention.}
  \label{fig:method_arch_diag}
\end{figure*}

Videos, like most time-series data, contain both stationary (information with constant statistical moments in time) and non-stationary components. State-of-the-art video learning networks, however, have not been designed with explicit learning of stationary and non-stationary information in mind. This, we believe, can inhibit robust forecasting 
\cite{liu2022non,10273738,zhou2022fedformer,wu2021autoformer}, which is a key factor for the tasks defined based on our dataset. To address this issue and effectively perform the tasks facilitated with CycleCrash, we design VidNeXt (see \cref{fig:arch_diag1}), a model that employs a dedicated spatial learner followed by temporal encoding of both stationary and non-stationary information. 

The spatial component of VidNeXt is a ConvNeXt~\cite{liu2022convnet} feature extractor. This encoder is a CNN which borrows design elements from transformers and is, therefore, more compatible with them \cite{liu2022convnet} through leveraging layer normalization and GELU activations \cite{hendrycks2016gaussian}. We empirically find that the base version of ConvNeXt \cite{liu2022convnet} works best for our purpose. This is followed by an encoder-decoder transformer \cite{liu2022non} comprising transformer blocks with de-stationary attention (see \cref{fig:arch_diag2}) which was originally designed for time-series forecasting.

The temporal component in VidNeXt, as demonstrated in \cite{liu2022non}, involves the normalization of a series of embeddings (see \cref{fig:arch_diag1}), along with the computation of statistical moments such as mean and variance. This normalization process results in stationary series $x'$ containing stationary information. Subsequently, the attention mechanism calculates attention scores based on this stationary series. To incorporate the non-stationary component into the attention scores, they are rescaled using re-scaling factors ($\Delta$) and ($\tau$) in the self-attention block, referred to as `de-stationary attention'. These factors are determined via a projector based on the raw data $x$ and the statistical moments calculated during normalization. Following this, the output embeddings from the encoder-decoder architecture ($y'$) are de-normalized using the previously calculated means and variances. Finally, an MLP head is applied to the video embeddings according to the specific requirements of one of a number of tasks.

The motivation behind disentangling complementary elements, namely stationary and non-stationary information, stems from two main factors. First, stationarization aims to attenuate non-stationarity, leading to improved predictability. Second, while stationarized series may offer better predictability, they may offer limited insights for forecasting real-world spurious events. This limitation can result in over-stationarization, where transformers generate indistinguishable temporal attention for different series. Thus, re-incorporating intrinsic non-stationary information such as statistical moments of spatial embeddings is crucial for enhancing overall predictability.
The non-stationary transformer in \cite{liu2022non} was originally used for time series forecasting with an input time series sequence of length 96 and a prediction sequence of length 48. We modify the time series sequence length to match the number of frames in the input, which in our case is 30.
We also set the prediction length to a value of 1, to indicate a 1D vector representing the whole video. 
We use the same output embedding dimension of ConvNeXt as the input embedding dimension length of the encoder and decoder.

\section{Experiments}\label{sec:experiments}
\noindent \textbf{Dataset Splits.}
We randomly partition the videos into training and testing sets with $7:3$ ratio. Each video is then segmented into 1-second sequences with 0.5-second overlaps. We provide the code to obtain the fixed splits to allow 
fair comparisons. 

\noindent \textbf{Baseline Models.}
\label{subsec:baselines}
For baseline comparisons, we have selected models widely adopted in the community for video processing tasks: ResNet50 3D~\cite{hara2018can}, R(2+1)D~\cite{tran2018closer}, X3D~\cite{feichtenhofer2020x3d}, TimeSformer~\cite{gberta_2021_ICML}, and  ViViT~\cite{arnab2021vivit}. ResNet50 3D, R(2+1)D, and X3D are used as common convolutional video learners. Notably, we use the X3D-XS, X3D-S, and X3D-M versions, which use 4, 13, and 16 input frames respectively. Additionally, we include TimeSformer and ViViT as competitive transformer-based methods. All the baseline models input a common frame size of $224 \times 224$.  In addition to the baselines, we present two \textbf{ablation variants} to isolate the impact of the spatial and temporal components of VidNeXt. In the first variant (ConvNeXt+VT), we use a vanilla transformer (VT) instead of a non-stationary transformer (NST) for the temporal component, while the second variant (ResNet+NST) involves ResNet-18~\cite{he2016deep} instead of ConvNeXt for the spatial component.

\noindent \textbf{Training Details.}
\label{subsec:training_proc}
We train all models on 30-frame (1 sec.) video clips. The networks are trained using batch size 32, AdamW~\cite{loshchilov2017decoupled} optimization, and  ReduceLROnPlateau scheduling with learning rate $2$e$^{-6}$ for 50 epochs.  For all the encoder backbones used in this study, including baseline models and components of our proposed method, we use the pretrained weights provided with their public distributions \cite{hara2018can,tran2018closer,feichtenhofer2020x3d,gberta_2021_ICML,arnab2021vivit,liu2022convnet,he2016deep}. 
As indicated earlier, all video frames are initially  1280 $\times$ 720 pixels.
Following \cite{Recasens_2021_ICCV,guo2022improving}, we first perform augmentation on all video sequences by randomly cropping at 700$ \times$ 700 pixels followed by random flipping with a probability of 0.25. 
The `Time-to-collision Prediction' task is inherently imbalanced because segmentation tends to produce most labels in videos with shorter time-to-collision values and fewer with longer ones. To address this, we upsampled videos by duplicating and augmenting brightness, contrast, saturation, and hue to the training data.
We use the spatial dimension of 224 $\times$ 224 pixels. We utilize cross-entropy loss for classification tasks and mean squared error for the regression tasks. Additionally, we perform mixed-precision training~\cite{micikevicius2017mixed} on two 40 GB NVIDIA A100 GPUs to save the computation overhead. We use the available implementations of the baseline models in PyTorch.

\begin{table*}[t]
\centering
\setlength\tabcolsep{5pt}
\resizebox{1\textwidth}{!}{
\begin{tabular}{lccccccccccccccccccc}
\toprule
\textbf{Method} &
\multicolumn{2}{c}{\textbf{Risk}} &
\multicolumn{2}{c}{\textbf{Right-of-way}} &
\multicolumn{2}{c}{\textbf{Collision}} &
\textbf{Time-to-coll.} &
\multicolumn{2}{c}{\textbf{Severity}} &
\multicolumn{2}{c}{\textbf{Fault}} &
\multicolumn{2}{c}{\textbf{Age}} &
\multicolumn{2}{c}{\textbf{Direction}} &
\multicolumn{2}{c}{\textbf{Object}} \\
\cmidrule(lr){2-3} \cmidrule(lr){4-5} \cmidrule(lr){6-7} \cmidrule(lr){8-8} \cmidrule(lr){9-10}
\cmidrule(lr){11-12} \cmidrule(lr){13-14} \cmidrule(lr){15-16} \cmidrule(lr){17-18}
&
\textbf{Acc.$\uparrow$} & \textbf{F1$\uparrow$} &
\textbf{Acc.$\uparrow$} & \textbf{F1$\uparrow$} &
\textbf{Acc.$\uparrow$} & \textbf{F1$\uparrow$} &
\textbf{MSE$\downarrow$} &
\textbf{Acc.$\uparrow$} & \textbf{F1$\uparrow$} &
\textbf{Acc.$\uparrow$} & \textbf{F1$\uparrow$} &
\textbf{Acc.$\uparrow$} & \textbf{F1$\uparrow$} &
\textbf{Acc.$\uparrow$} & \textbf{F1$\uparrow$} &
\textbf{Acc.$\uparrow$} & \textbf{F1$\uparrow$} \\
\midrule
TimeSformer~\cite{gberta_2021_ICML} & 65.74 & \underline{41.79} & 60.20 & 55.71 & \underline{66.45} & 69.69 & 1.41 & 36.49 & 23.01 & 59.65 & 51.03 & 93.77 & 66.68 & 47.19 & \underline{31.38} & \underline{45.02} & \textbf{29.00} \\
ViViT~\cite{arnab2021vivit} & 65.12 & 39.06 & 52.84 & 53.74 & 57.01 & \underline{69.92} & \textbf{1.33} & 47.51 & 24.47 & 53.37 & 50.42 & 93.56 & 66.34 & 36.29 & 27.99 & \textbf{46.30} & 26.34 \\
ResNet50 3D~\cite{hara2018can} & 65.76 & 39.53  & 59.41 & 53.97 & 63.10 & 60.24 & \underline{1.38} & 56.60 & 26.12 & 59.37 & \underline{54.91} & 94.21 & 54.86 & 46.30 & 30.12 & 43.27 & 27.77 \\
R(2+1)D~\cite{tran2018closer} & \underline{66.54} & 39.56 & 60.31 & 53.42 & \textbf{67.71} & 63.33 & 1.43 & \underline{56.63} & 25.46 & 50.53 & 52.62 & 94.41 & 53.24 & 47.49 & 30.36 & 40.75 & 25.48 \\
X3D-M~\cite{feichtenhofer2020x3d} & 64.76 & 38.75 & 59.83 & \underline{57.57} & 63.72 & 61.08 & 1.44 & 54.45 & 24.70 & 52.16 & 52.19 & 94.34 & 53.78 & \underline{47.82} & 31.85 & 42.72 & 23.79 \\
X3D-S~\cite{feichtenhofer2020x3d} & 63.37 & 36.28 & 60.10 & 56.90 & 61.49 & 61.13 & 1.47 & 51.80 & 24.09 & \underline{60.47} & 51.88 & 94.38 & 54.30 & 45.62 & 30.05 & 42.03 & 22.38 \\
X3D-XS~\cite{feichtenhofer2020x3d} & 64.77 & 37.23 & 59.37 & 53.43 & 60.59 & 60.73 & 1.47 & 51.39 & 23.77 & 56.10 & 52.59 & 93.87 & 52.37 & 46.77 & 30.22 & 41.73 & 26.57 \\
\hline
ConvNeXt+VT & 64.89 & 40.05 & 61.13 & 54.00 & 63.50 & 65.06 & 1.56 & 53.80 & \underline{26.54} & 56.74 & \textbf{55.72} & \underline{94.55} & \underline{66.78} & 46.46 & \textbf{32.62} & 42.85 & 25.16 \\
ResNet+NST & \textbf{67.18} & 40.74 & \underline{61.77} & \textbf{58.62} & 60.79 & 62.28 & 1.39 & 53.88 & 24.67 & 57.17 & 54.08 & 94.52 & 53.49 & 45.12 & 28.48 & 44.17 & 26.91 \\
\textbf{VidNeXt (Ours)} & 66.20 & \textbf{41.96} & \textbf{64.28} & 57.51 & 64.84 & \textbf{70.84} & \underline{1.38} & \textbf{59.66} & \textbf{31.78} & \textbf{65.16} & 52.51 & \textbf{94.57} & \textbf{67.88} & \textbf{47.94} & 31.20 & 42.31 & \underline{28.37}\\
\bottomrule
\end{tabular}
}
\caption{Combined experimental results for tasks 1 through 9. The methods above the line are baselines based on prior works, while those below the line are VidNeXt and its \textbf{ablation} variants.}
\label{tab:experimental_results_combined}
\end{table*}

\noindent \textbf{Results.}
\label{subsec:results} 
We perform the 9 tasks described earlier in \cref{subsec:tasks_and_metrics} and present the results in \cref{tab:experimental_results_combined}. We observe that for the majority of tasks defined in CycleCrash, VidNeXt outperforms prior methods. This is evident as the proposed method outperforms the others on most tasks,  
for instance `Cyclist Behavior Risk Index Prediction', `Right-of-way Classification', `Cyclist Collision Anticipation', `Severity Classification', `Fault Classification', `Cyclist Age Classification', and `Cyclist Direction Detection'. Among two tasks where VidNeXt does not perform the best, it nevertheless yields the second-best performance in metrics such as MSE in `Time-to-collision Prediction' and F1 in `Direction of the Object Involved Detection'. We notice a clear difference between the performance of the ablation variants and the proposed method in almost every task, with the highest margin of improvement being up to 8.4\%
in accuracy and 14.4\% of F1. 
We also observe that among the other baselines and ablation variants, ConvNeXt+VT and ResNet+NST generally emerge as the second-best or third-best models following VidNeXt.

\begin{table}[t]
\centering

\setlength\tabcolsep{2pt} 
\resizebox{1\columnwidth}{!}{
\begin{tabular}{lllllll}
\hline
\textbf{Trained on} & \textbf{Vehicles} & \textbf{Splits} & \textbf{Tested on} & \textbf{Eval.} & \textbf{Accuracy} & \textbf{F1} \\
\hline
\multirow{3}{4em}{CCD~\cite{10.1145/3394171.3413827}}
& \multirow{3}{*}{Cars} & \multirow{3}{*}{1} & CCD & - & 76.14 & 66.23 \\
\cline{4-7}
& & & CycleCrash & Finetuning & 55.76 & 25.28 \\
& & & CycleCrash & Linear & 59.78 $(+4.0)$ & 44.28 $(+21.5)$ \\
\hline
\multirow{3}{4em}{DoTA~\cite{yao2020and}}
& \multirow{3}{*}{Cars, Cyclist} & \multirow{3}{*}{3} & DoTA & - & 77.61 & 84.03 \\
\cline{4-7}
& & & CycleCrash & Finetuning & 56.57 & 60.75 \\
& & & CycleCrash & Linear & 59.51 $(+2.9)$ & 62.12 $(+1.4)$ \\
\hline
\multirow{5}{4em}{CycleCrash (Ours)}
& \multirow{5}{*}{Cyclist} & \multirow{5}{*}{1} & CycleCrash & - & 67.71 & 63.33 \\
\cline{4-7}
& & & CCD & Finetuning & 54.81 & 54.17 \\
& & & CCD & Linear & 62.44 $(+7.6)$ & 53.28 $(-0.9)$ \\
\cline{4-7}
& & & DoTA & Finetuning & 63.62 & 73.14 \\
& & & DoTA & Linear & 68.78 $(+5.2)$ & 81.22 $(+8.1)$ \\
\hline
\end{tabular}
}
\caption{Results of cross-dataset evaluation for collision anticipation using R(2+1)D.}
\label{tab:cross_dataset_results}
\end{table}


We perform cross-dataset evaluations on the `Collision Anticipation' task using  CCD~\cite{10.1145/3394171.3413827}, DoTA~\cite{yao2020and}, and CycleCrash datasets to investigate the transferability of learned representations. 
Both CCD and predominantly involve only car collision events, and so the extent to which collisions with cars and collisions with cyclists share visual cues remains to be explored. We opt for the R(2+1)D model due to its strong performance in this specific task and perform both finetuning(updates all model parameters) and linear evaluations (only updates the final layer, keeping the rest of the model frozen). The results of these evaluations are presented in \cref{tab:cross_dataset_results}. 
We observe from this table that training on CycleCrash and testing on CCD and DoTA yields considerably better results than the other way around, indicating that CycleCrash contains more information and situations regarding collisions, filling a unique gap in the landscape of datasets in this area. This experiment highlights CycleCrash's potential as a rich resource for pretraining models focused on collisions and vehicle safety

\begin{figure}
  \centering
  \includegraphics[width=0.6\linewidth]{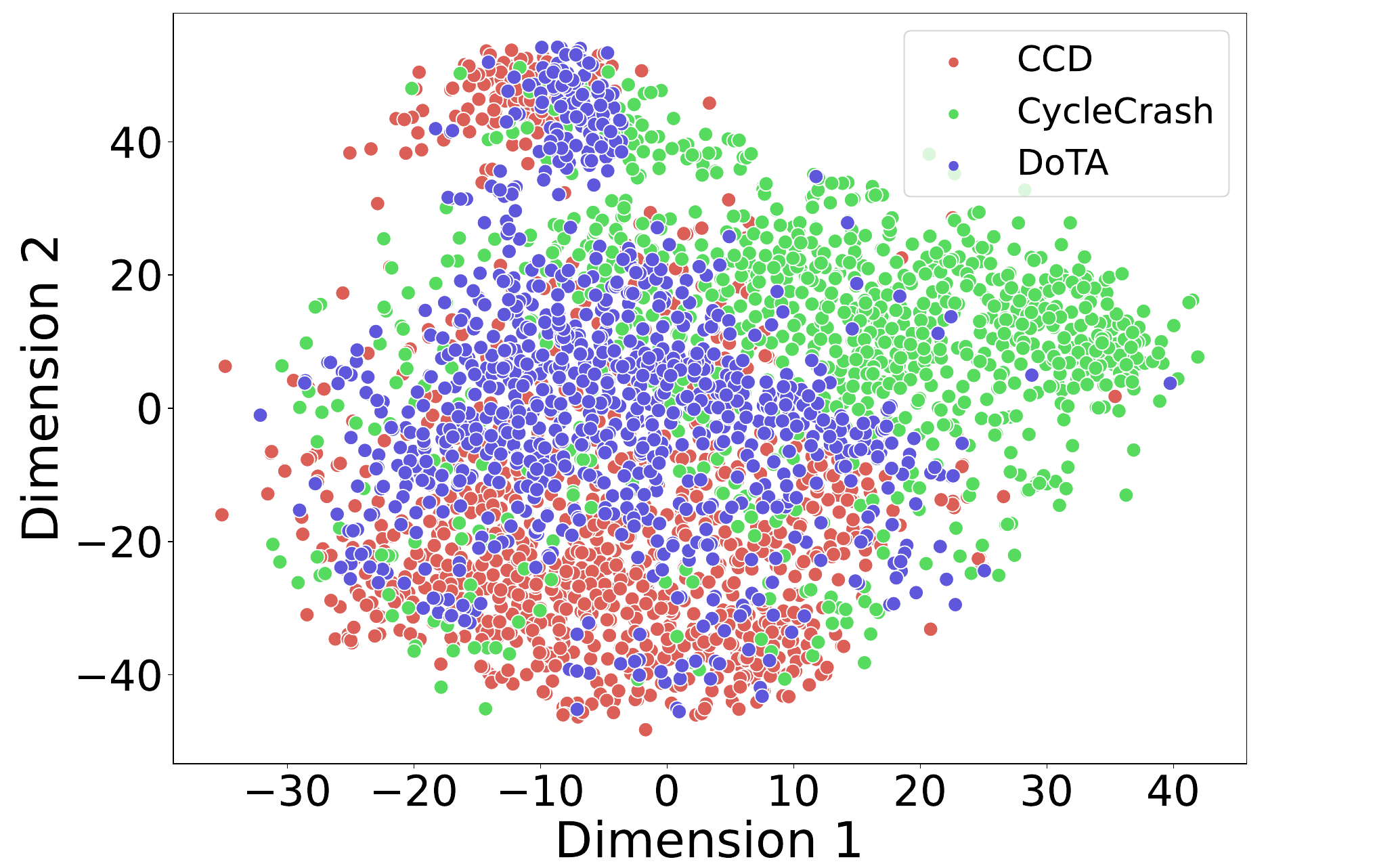}
  \caption{t-SNE plot of representations from CycleCrash and two other datasets.
  }
  \label{fig:visualize_datasets}
\end{figure}

We evaluate the distribution of the embeddings of our dataset with respect to CCD and DoTA, in~\cref{fig:visualize_datasets} we visualize 1,000 random samples from each of the datasets using t-SNE. We use a standard Kinetics400~\cite{kay2017kinetics} pre-trained model for this purpose. We observe in this figure that CCD and DoTA datasets considerably overlap with one another. However, while both CCD and DoTA overlap with CycleCrash, our dataset occupies a wider area of the latent space, indicating that it contains information not present in either CCD or DoTA. This finding highlights the diverse and distinct scenarios provided in our dataset.

\section{Conclusion}
In this paper, we present the CycleCrash dataset, a unique collection of 3,000 video clips capturing cyclists in various real-world scenarios, with a particular emphasis on potential cyclist safety in urban settings. To the best of our knowledge, CycleCrash is the first dataset to focus on cyclist-based collision prevention and understanding. The dataset includes 13 annotations for collision, cyclist, and scene-related labels, and defines 9 collision prevention and mitigation tasks. Additionally, we propose VidNeXt, a novel method for video classification and regression tailored to learn both stationary and non-stationary information. Furthermore, we evaluate the proposed VidNeXt method against seven baselines and ablation variants on CycleCrash, to demonstrate its stronger performance.

\noindent \textbf{Acknowledgment.} We would like to thank Geotab Inc., the City of Kingston, and NSERC for supporting this work, as well as Kate Cowperthwaite, Debaditya Shome, and the annotators who contributed to this project.

\clearpage
{\small
\bibliographystyle{ieee_fullname}
\bibliography{egbib}
}

\clearpage

\appendix
\section*{\LARGE Appendix}

\section*{A. Dataset Availability}\label{sec:availability}
The dataset is publicly available and can be accessed at \url{https://github.com/DeSinister/CycleCrash/}. The 
initial release of the dataset comprises:
\begin{enumerate}[label=\textbullet, nosep]
    \item List of links for all the videos in the dataset, with their start and end timestamps.
    \item All 13 annotations described in the paper, for all the video clips.
    \item PyTorch-based implementation for the pre-processing used in the paper, and data loader codes for efficiency and consistency.
    \item PyTorch-based implementation for VidNeXt and its ablation variants.
\end{enumerate}

\section*{B. Keyword Search for dataset creation}\label{sec:keywords}
For creating the dataset, the following keywords were used to search and collect the videos in CycleCrash: 
\texttt{Bicycle accident dashcam; Bicycle crash; Bicycle crash dashcam; Bicyclist car accident; 
Bicyclist car collision dashcam; Bicyclist collision; Bicyclist collision compilation; 
Bicyclist crash; Bike accident; Bike crash; Bike crash compilation; Bike hit car dashcam; 
Bike hit dashcam; Bike hit and run; Car hit cyclist; Cyclist accident; Cyclist accident dashcam; 
Cyclist being hit; Cyclist being hit compilation; Cyclist collision dashcam; Cyclist crash dashcam; 
Cyclist fall down; Cyclist fault dashcam; Cyclist hit and run; Cyclist hit dashcam; Cyclist hit pedestrian; 
Cyclist hits bus; Cyclist hits pole; Cyclist hitting car; Cyclist hits road sign; Cyclist hits train; 
Cyclist near miss; Dashcam cycle accident; Dashcam cycle accident compilation; Hit cyclist.}

For the videos exhibiting cyclists navigating safely in an urban traffic scenario, without depicting risky behaviour, we use the clips from driving tour videos collected using the following search keywords:
\texttt{City Driving tour;
Downtown Driving tour;
Urban Driving tour;
Urban City Driving tour.}

\section*{C. Multi-tasking}\label{sec:multitask}
We performed a multi-task version of our model VidNeXT and present the results below in \cref{tab:multi_task}. The multitasking model slightly underperforms possibly due to task interference.

\begin{table*}[t]
\centering

\setlength\tabcolsep{5pt}
\resizebox{1\textwidth}{!}{
\begin{tabular}{lccccccccccccccccccc}
\toprule
\textbf{Method} &
\multicolumn{2}{c}{\textbf{Risk}} &
\multicolumn{2}{c}{\textbf{Right-of-way}} &
\multicolumn{2}{c}{\textbf{Collision}} &
\textbf{Time-to-coll.} &
\multicolumn{2}{c}{\textbf{Severity}} &
\multicolumn{2}{c}{\textbf{Fault}} &
\multicolumn{2}{c}{\textbf{Age}} &
\multicolumn{2}{c}{\textbf{Direction}} &
\multicolumn{2}{c}{\textbf{Object}} \\
\cmidrule(lr){2-3} \cmidrule(lr){4-5} \cmidrule(lr){6-7} \cmidrule(lr){8-8} \cmidrule(lr){9-10}
\cmidrule(lr){11-12} \cmidrule(lr){13-14} \cmidrule(lr){15-16} \cmidrule(lr){17-18}
&
\textbf{Acc.$\uparrow$} & \textbf{F1$\uparrow$} &
\textbf{Acc.$\uparrow$} & \textbf{F1$\uparrow$} &
\textbf{Acc.$\uparrow$} & \textbf{F1$\uparrow$} &
\textbf{MSE$\downarrow$} &
\textbf{Acc.$\uparrow$} & \textbf{F1$\uparrow$} &
\textbf{Acc.$\uparrow$} & \textbf{F1$\uparrow$} &
\textbf{Acc.$\uparrow$} & \textbf{F1$\uparrow$} &
\textbf{Acc.$\uparrow$} & \textbf{F1$\uparrow$} &
\textbf{Acc.$\uparrow$} & \textbf{F1$\uparrow$} \\
\midrule
ViViT~\cite{arnab2021vivit} (single-task) & 65.12 & 39.06 & 52.84 & 53.74 & 57.01 & \underline{69.92} & \textbf{1.33} & 47.51 & 24.47 & 53.37 & 50.42 & 93.56 & 66.34 & 36.29 & 27.99 & \textbf{46.30} & 26.34 \\
X3D-M~\cite{feichtenhofer2020x3d} (single-task) & 64.76 & 38.75 & 59.83 & \underline{57.57} & 63.72 & 61.08 & 1.44 & 54.45 & 24.70 & 52.16 & 52.19 & 94.34 & 53.78 & \underline{47.82} & 31.85 & 42.72 & 23.79 \\
\textbf{VidNeXt (single-task)} & 66.20 & \textbf{41.96} & \textbf{64.28} & 57.51 & 64.84 & \textbf{70.84} & \underline{1.38} & \textbf{59.66} & \textbf{31.78} & \textbf{65.16} & 52.51 & \textbf{94.57} & \textbf{67.88} & \textbf{47.94} & 31.20 & 42.31 & \underline{28.37}\\
\textbf{VidNeXt (multi-task)} & 51.86 & 31.34 & 59.09 & 55.38 & 62.86 & 62.87 & 2.55 & 45.17 & 29.62 & 57.16 & 51.43 & 93.56 & 65.66 & 44.85 & 31.04 & 40.4 & 27.64\\
\bottomrule
\end{tabular}
}
\caption{Multi-task version of VidNeXT. }
\label{tab:multi_task}
\end{table*}

\begin{figure}[htbp]
        \centering
            \begin{subfigure}{\linewidth}
            \centering
            \includegraphics[width=\linewidth]{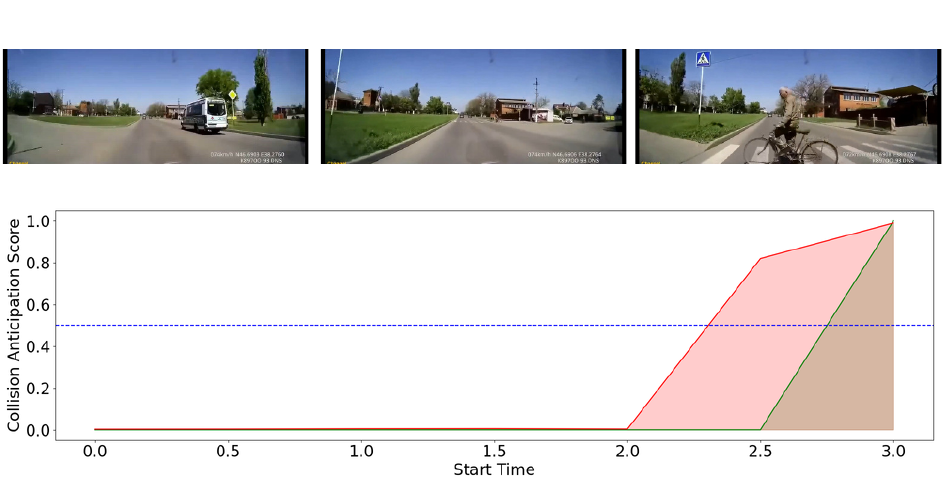}
            \caption{Failure Case 1}
            \end{subfigure}
        
            \begin{subfigure}{\linewidth}
            \centering
            \includegraphics[width=\linewidth]{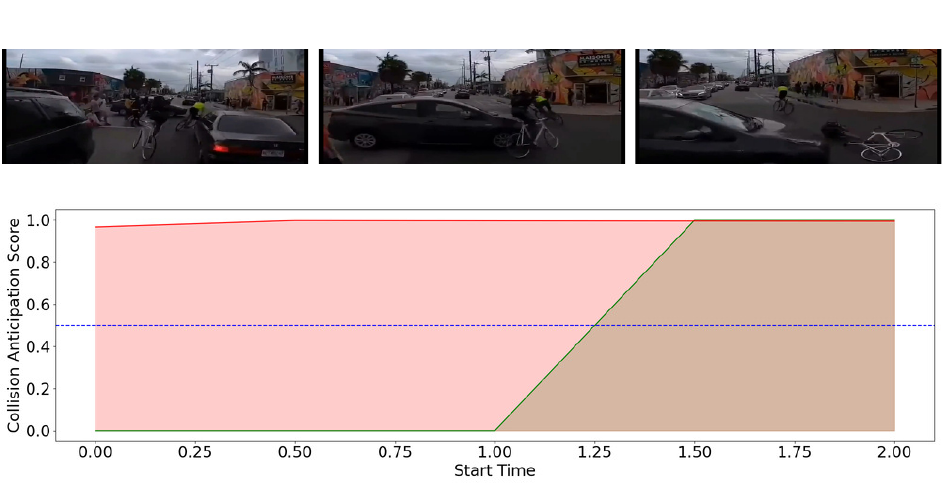}
            \caption{Failure Case 2}
            \end{subfigure}

            \begin{subfigure}{\linewidth}
            \centering
            \includegraphics[width=\linewidth]{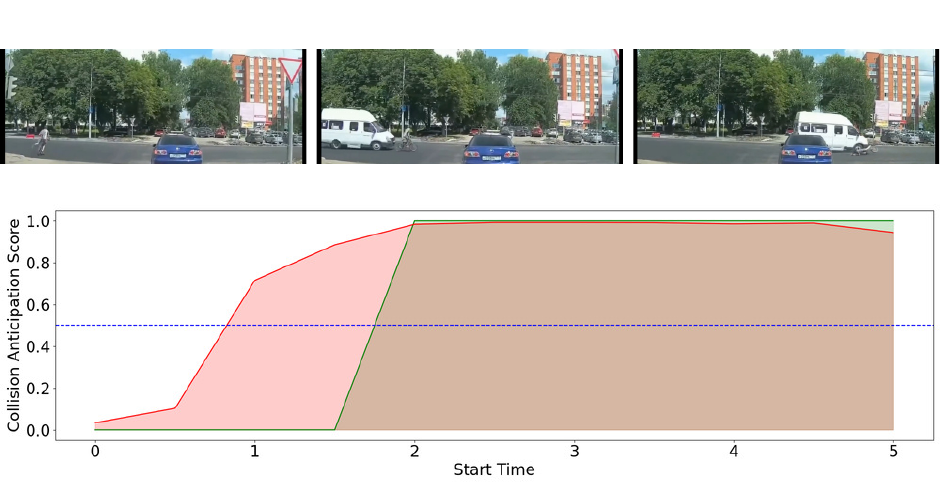}
            \caption{Failure Case 3}
            \end{subfigure}
    \caption{A few examples of failure cases of VidNeXt for the Collision Anticipation task. The red line represents the collision anticipation scores predicted by the model, while the green line shows the ground truth labels. The blue line marks the threshold at a score of 0.5, which is used to determine whether a collision is anticipated or not.} 
    \label{fig:failure_cases}
\end{figure}

\section*{D. Future Research Directions}

We visualize a few examples of the failure cases of our method for the Collision Anticipation task, in \cref{fig:failure_cases}. We observe that in these instances, the model predicts an accident earlier than it actually occurs. Investigating the underlying reasons for this and designing effective solutions are an interesting line of future inquiry.
Additionally to enhance overall training and performance, specialized losses, long-term recognition methods, or spatiotemporal augmentation techniques could be employed to upsample the minority classes. 
Moreover, incorporating scene-related labels such as `type of object involved,' `camera position,' and `ego-vehicle involved' could provide additional context to improve model performance. Finally, we have introduced tasks such as `Right-of-way Prediction' and `Severity Classification', which have the potential to be extended to include interactions among autonomous vehicles and other vehicles (cars, motorcycles, buses, etc).

\section*{E. Additional Data Statistics}\label{sec:additional_data_stats}

\begin{figure}[ht]
  \centering
   \includegraphics[width={\linewidth}]{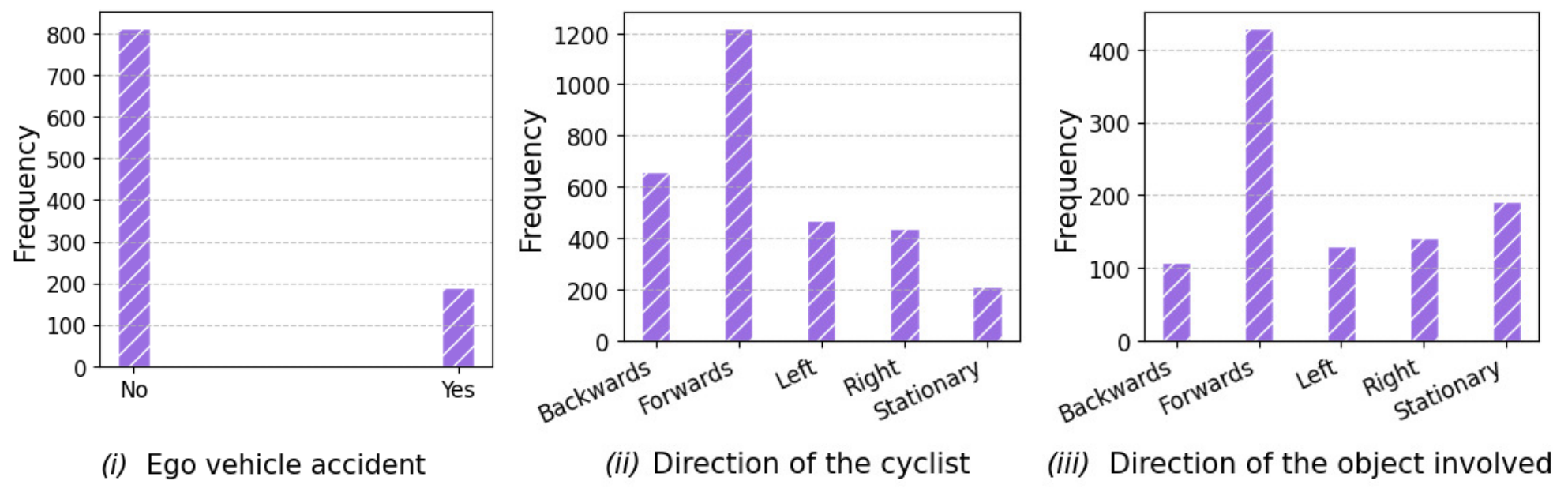}

 \caption{Distribution of CycleCrash data for (\textit{i}) ego vehicle accidents, (\textit{ii}) direction of cyclists, (\textit{iii}) direction of other objects involved. 
    }
  \label{fig:data_hist2}
\end{figure}
  
\begin{figure}[ht]
    \includegraphics[width=\linewidth]{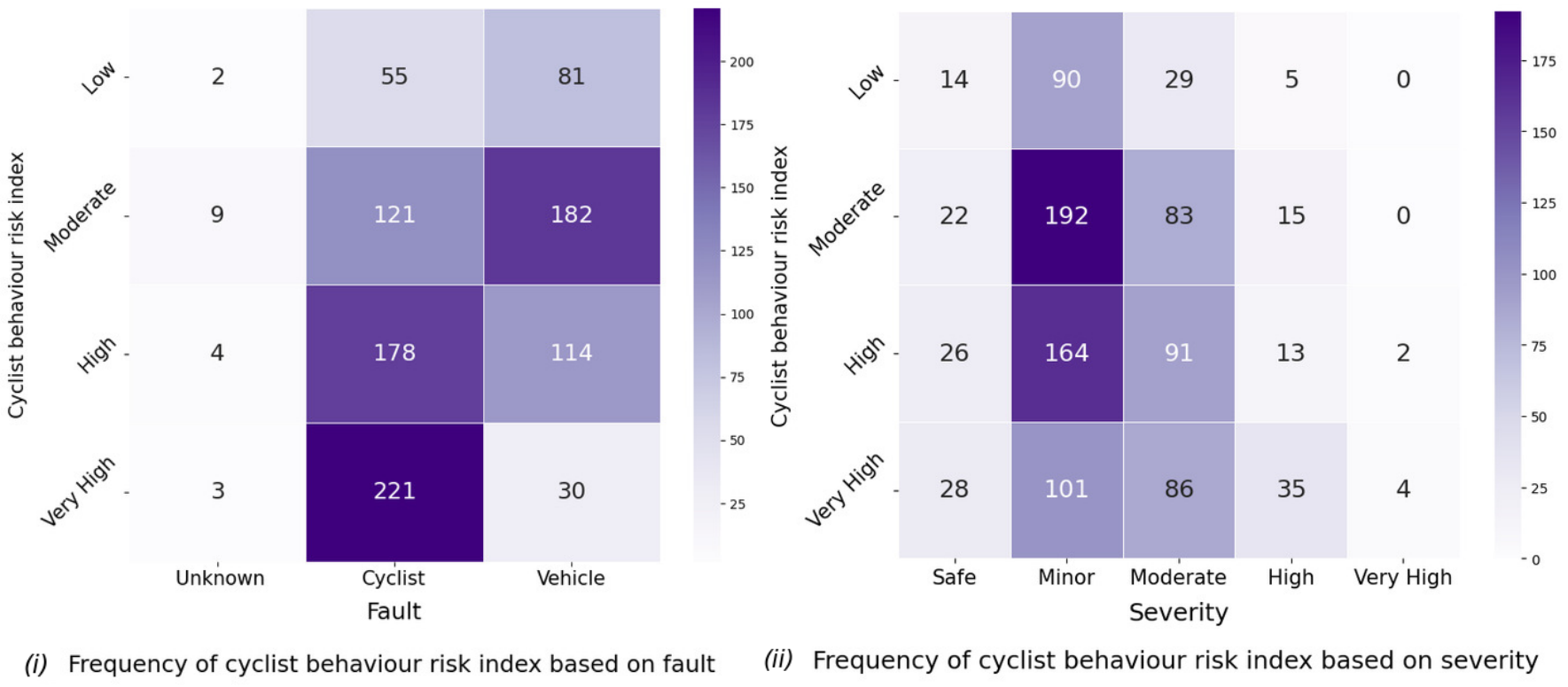}
  \caption{ Relationship between (\textit{i}) cyclist behaviour risk index and fault, (\textit{ii}) cyclist behaviour risk index and severity.}
    \label{fig:data_heat2}
\end{figure}

We present more statistics for our dataset in \cref{fig:data_hist2} and \cref{fig:data_heat2}. Here, \cref{fig:data_hist2} (\textit{i}) indicates if the ego-vehicle (the vehicle carrying the dashcam) is involved in the collision in the video. We observe that most of the videos do not involve the ego-vehicle in the accident. Furthermore, we present the distributions of the directions of the cyclists in \cref{fig:data_hist2} (\textit{ii}), and the directions of the objects involved in \cref{fig:data_hist2} (\textit{iii}). We observe that the most frequently occurring direction was forward in both cyclists and other objects involved. \cref{fig:data_heat2} (\textit{i}) displays the frequency of fault based on the cyclist behaviour risk index. It is observed that as the cyclist behaviour risk index increases, the number of cyclists being at fault also rises, and vice-versa. Moreover, the relationship between cyclist behaviour risk index and severity is presented in \cref{fig:data_heat2} (\textit{ii}). We notice an increasing number of collisions in each severity group as the cyclist behaviour risk index increases. The highest number of collisions was recorded with minor collision severity and low cyclist behaviour risk indexes.

\section*{F. Additional Visualizations}\label{sec:additional_viz}
\begin{figure*}[!h]
  \centering
    \includegraphics[width={0.82\linewidth}]{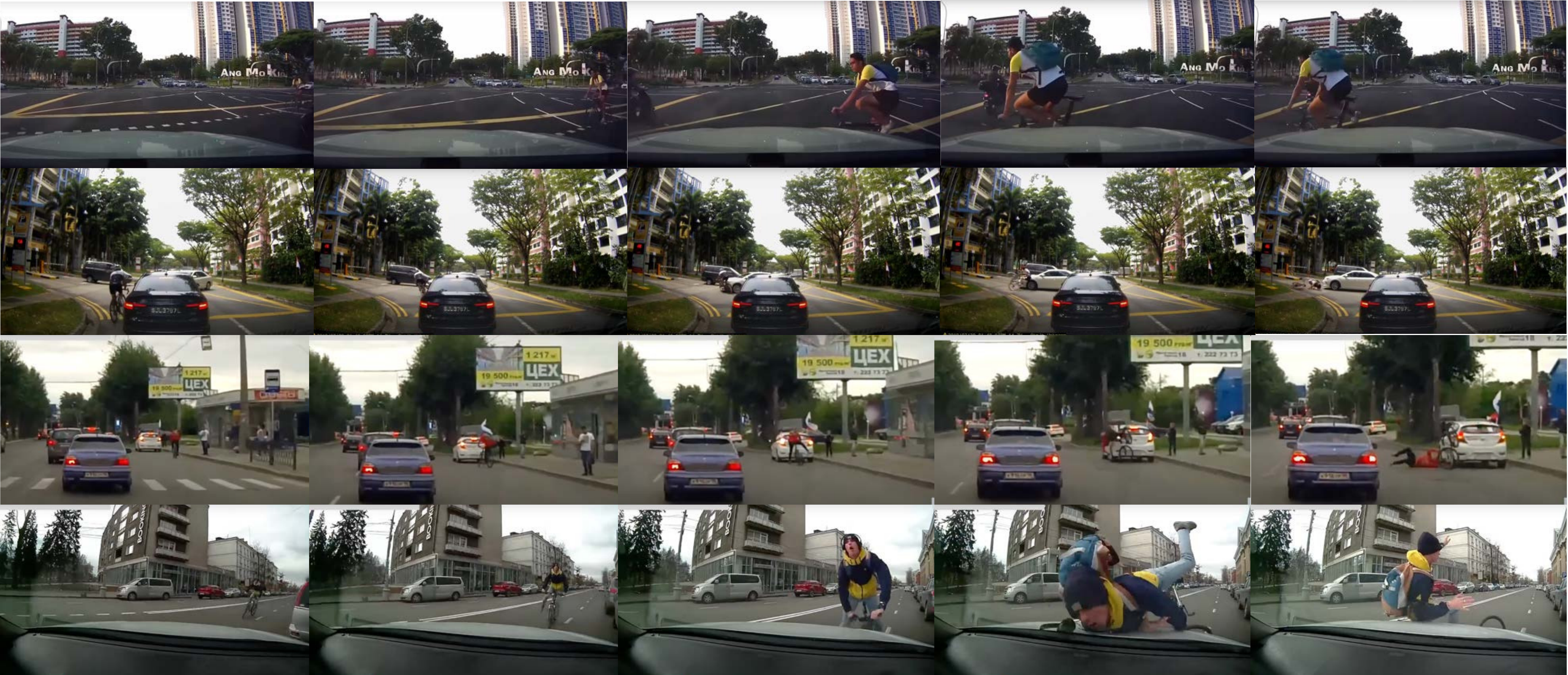}
  \caption{Additional visualizations of cyclist interactions with cars from the CycleCrash dataset.}
  \label{fig:snapshots_of_datasets}
\end{figure*}

\begin{figure*}[!h]
  \centering
    \includegraphics[width={0.82\linewidth}]{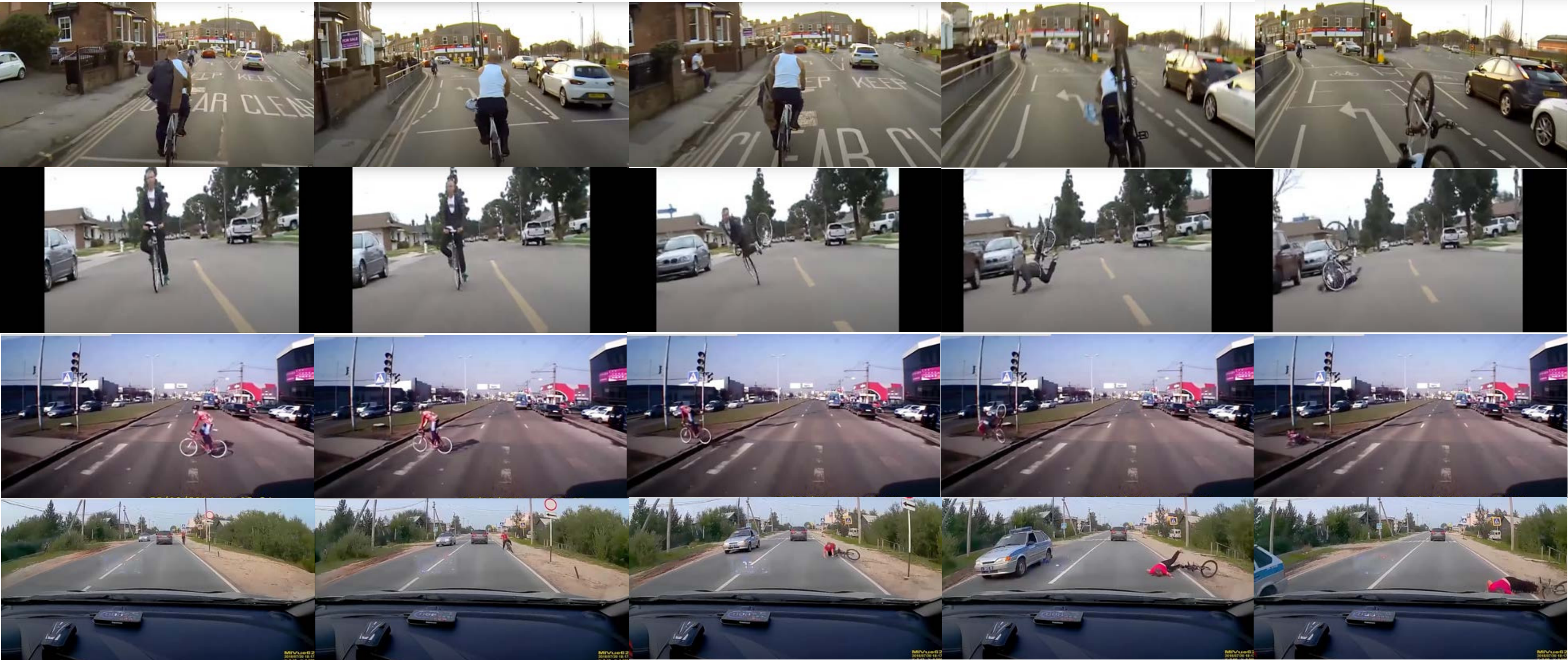}
  \caption{Additional visualizations of cyclists falling on their own from the CycleCrash dataset.}
  \label{fig:snapshots_of_datasets2}
\end{figure*}

\begin{figure*}[!h]
  \centering
    \includegraphics[width={0.82\linewidth}]{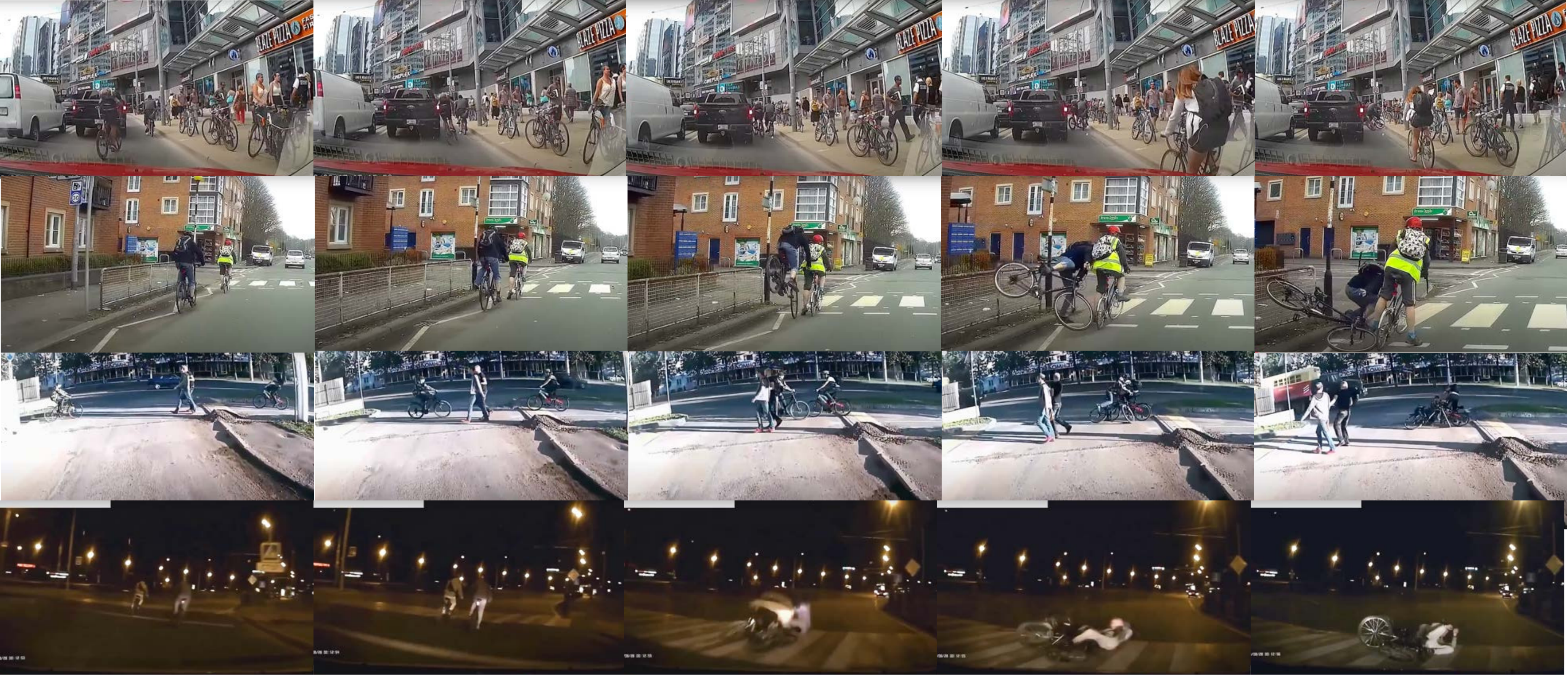}
  \caption{Additional visualizations of cyclist interactions with other cyclists from the CycleCrash dataset.}
  \label{fig:snapshots_of_datasets3}
\end{figure*}

\begin{figure*}[!t]
  \centering
  \vspace{-6in}
    \includegraphics[width={0.82\linewidth}]{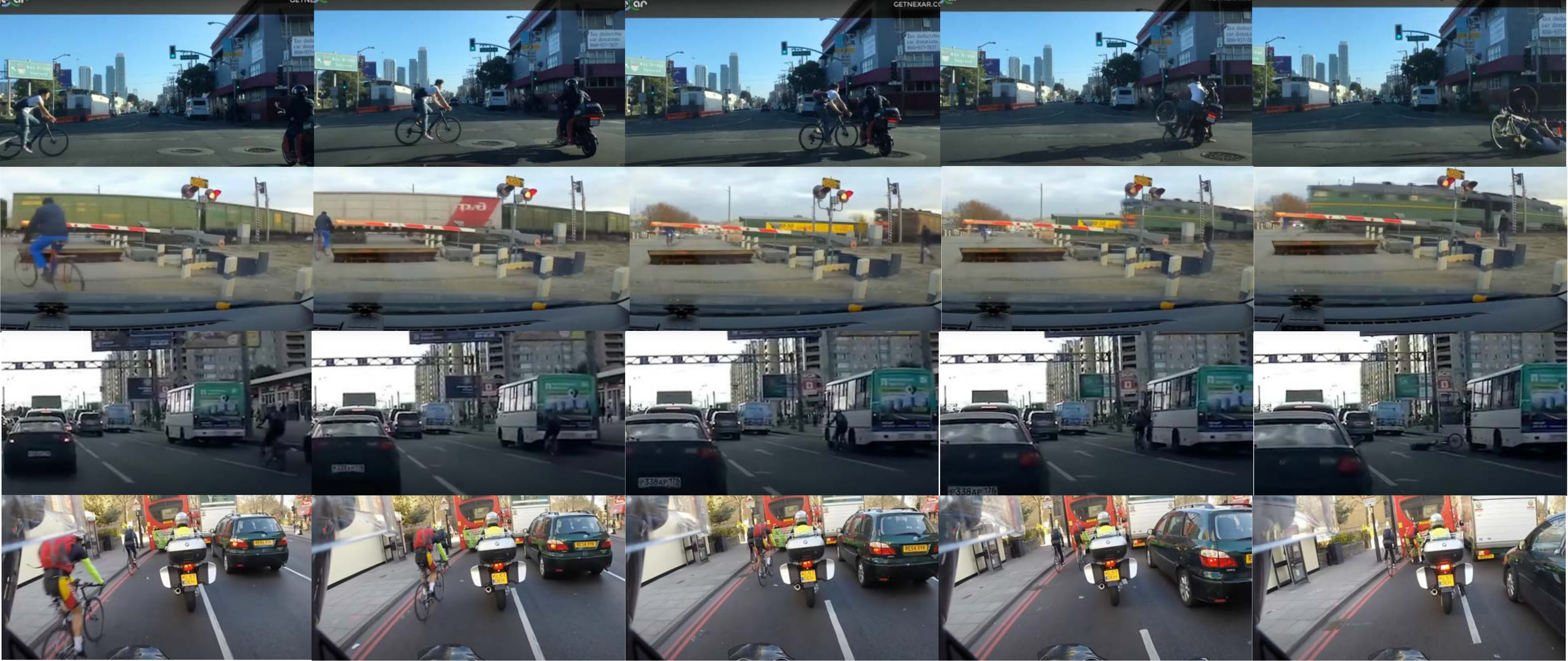}
  \caption{Additional visualizations of cyclist interactions with other motor vehicles from the CycleCrash dataset.}
  \label{fig:snapshots_of_datasets4}
\end{figure*}

We present additional samples for different video streams, showing diversity in interactions in the CycleCrash dataset in \cref{fig:snapshots_of_datasets} to \cref{fig:snapshots_of_datasets4}. 

\clearpage

\end{document}